\documentclass[runningheads]{llncs}

\usepackage{float}
\usepackage{algorithm}
\usepackage{algpseudocode}
\usepackage{tabularx}
\usepackage{array}
\newcolumntype{Y}{>{\raggedright\arraybackslash}X}

\makeatletter
\let\titleold\title
\renewcommand{\title}[1]{%
  \titleold{#1}%
  \gdef\thetitle{#1}%
}
\providecommand{\thetitle}{}

\newcommand{\maketitlesupplementary}{%
  \newpage
  \begin{center}
    \Large\textbf{\thetitle}\\[0.5em]
    Supplementary Material\\[1em]
  \end{center}
}
\makeatother
\usepackage{svg}

\usepackage{adjustbox}
\usepackage{eccvabbrv}

\usepackage{graphicx}
\usepackage{booktabs}

\usepackage[accsupp]{axessibility}  % Improves PDF readability for those with disabilities.

\usepackage[
    colorlinks=true,
    citecolor=blue,
    linkcolor=blue,
    urlcolor=blue
]{hyperref}

\usepackage{orcidlink}

\begin{document}

\title{HASSL: Hierarchy-Aware Self-Supervised Learning Framework for Single Cell Microscopy}

\titlerunning{HASSL}
\newcommand{\equalcontrib}{\textsuperscript{\textdagger}}

\author{Julius Riel\equalcontrib\inst{1,2,3}%\orcidlink{0000-1111-2222-3333}
\and
Vishwa Mohan Singh\equalcontrib\inst{1,2}% \orcidlink{1111-2222-3333-4444}
\and
Sai Anirudh Aryasomayajula\inst{1,2}%\orcidlink{2222--3333-4444-5555}
\and
Anuun Chinbat \inst{1} \and
Hannes Leonhard \inst{1,3} \and
Moritz Ladenburger \inst{1,3} \and
Frederik Alexander \inst{1,3} \and
Vishisht Choudhary \inst{1,3} \and
Fabio Laredo \inst{4} \and
Giacomo Masserdotti \inst{4} \and
Thorben Prein \inst{1,3} \and
Carsten Marr\textsuperscript{*} \inst{5,6,7,8,9} \and
Amirhossein Kardoost\textsuperscript{*} \inst{5,9}
}

\authorrunning{J.~Riel and V.~M.~Singh et al.}

%\institute{Princeton University, Princeton NJ 08544, USA \and
%Springer Heidelberg, Tiergartenstr.~17, 69121 Heidelberg, Germany

\institute{TUM.ai, Munich, Germany \and
Ludwig-Maximilian-University, Munich, Germany \and
Technical University of Munich, Munich, Germany \and
Institute of Stem Cell Research, Helmholtz Munich, German Research Center for Environmental Health, Neuherberg, Germany \and
Computational Health Center, Helmholtz Munich - German Research Center for Environmental Health, Neuherberg, Germany \and
Department of Medicine III, Ludwig-Maximilian-University Hospital, Munich, Germany \and
Department of Physics, Ludwig-Maximilian-University, Munich, Germany \and
German Cancer Consortium (DKTK), partner site Munich, Germany \and
Munich Center for Machine Learning (MCML), Munich, Germany
}

\maketitle

\begingroup
\renewcommand{\thefootnote}{\textdagger}
\footnotetext{Equal contribution.}

\renewcommand{\thefootnote}{\textasteriskcentered}
\footnotetext{Corresponding authors:
\email{\{carsten.marr,amirhossein.kardoost\}@helmholtz-munich.de}.}
\endgroup
\begin{abstract}
 Hierarchical structure is common in image data, where fine-grained clusters often merge into coarser semantic groups. In biological cell images, however, current self-supervised models often suppress this structure because coarse factors, such as imaging modality, dominate the latent space and obscure finer morphological attributes. We propose a hierarchy-aware self-supervised framework that preserves biologically meaningful structure in cellular representations. Our method combines two components: a segmentation-guided distillation teacher that improves morphological awareness, and an HDBSCAN-based hierarchy-aware contrastive loss that sharpens boundaries between related but distinct subtypes at each hierarchical level. Together, these components align embeddings with semantic and morphological cues while reducing the dominance of coarse acquisition factors. We train and evaluate our method on a curated corpus of 2.3M single cells from 20 microscopy datasets, covering 208 cell classes. Our approach improves over baseline methods, increasing average top-$K$ accuracy by $+2.8\%$, top-$9$ retrieval on deeply hierarchical data by $+6.3\%$, and downstream drug-classification F1-score from perturbed cell morphology by $+7.8\%$.

  %\todo{comment (Giuseppe, storytelling): Emphasize the value of the increase in numbers: e.g. from Table 1: avg top-K-Accuracy up by +2.78\% - "Accuracy K=9: 83.5\% (ours) vs 80.9\% (baseline)" - that shows a solid improvement since it's not just "40\% vs 43\%", but a higher-value increase) - same potentially with the other values, whatever appears best\n\n}

 % \todo{it is medium to high AI-generated text. please rephrase the abstract.}
 % \todo{Giuseppe (Spelling) - [Abstract, l. 28]: should be "baseline" instead of "basline"}
  

\end{abstract}
\noindent\textbf{Version note.}
This manuscript is a pre-peer-review preprint. The final published version may differ from this version.

\section{Introduction}
\label{sec:intro}

\begin{figure}[!b]
    \centering
    \includegraphics[width=\linewidth]{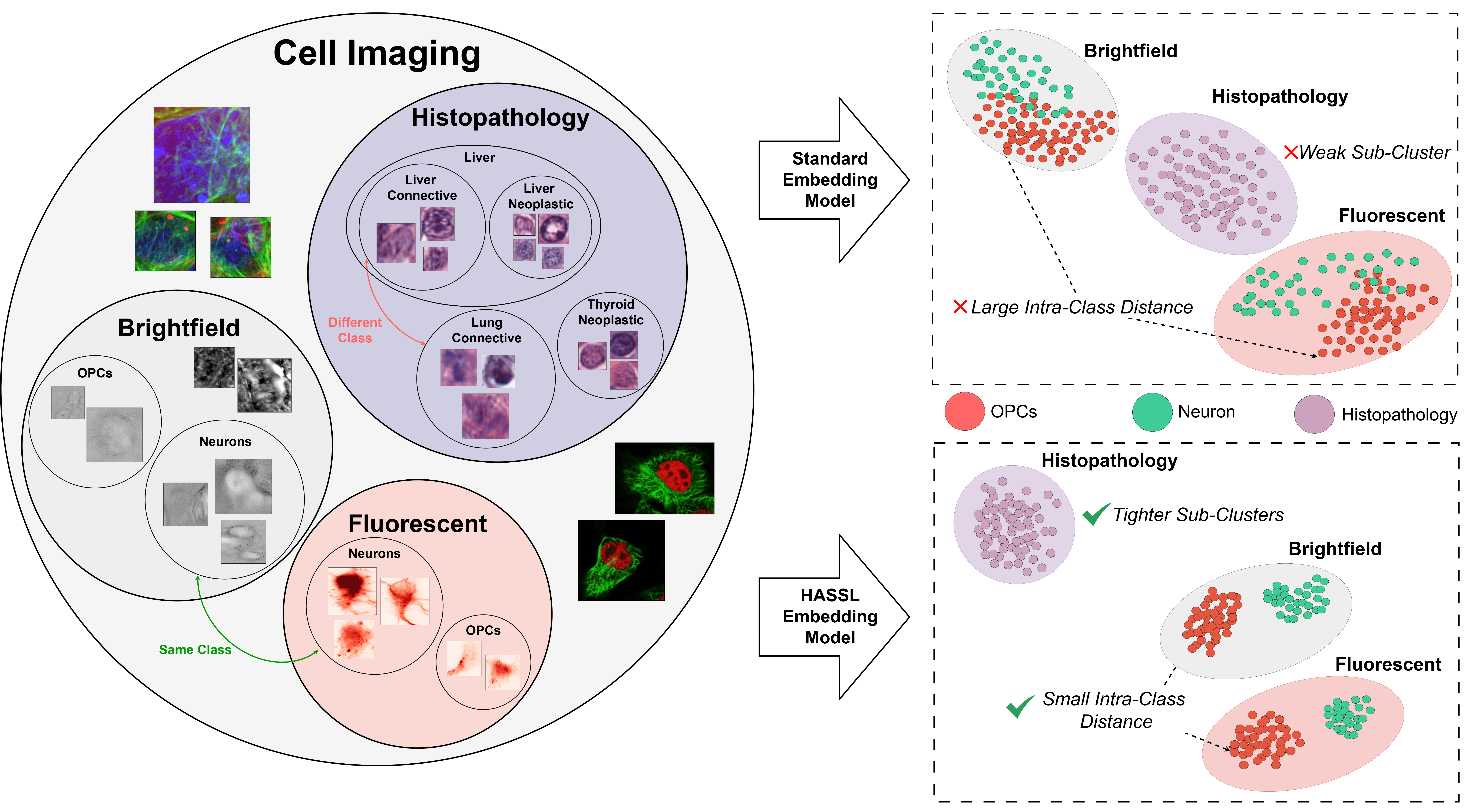}
    \caption{Hierarchy in cell imaging (left) and the objective with HASSL's embedding model (right). HASSL learns a more hierarchical embedding space, leading to improved morphological representation and tighter subclusters.}
    \label{fig:hierarchy}
\end{figure}

Accurately labeled microscopy datasets are scarce, motivating self-supervised learning (SSL)-based methods to encode biologically meaningful signals for downstream tasks~\cite{Kobayashi2022,kim2025self}. Distillation-based  models such as DINO~\cite{dino} scale well without labels and avoid contrastive large-batch or memory-bank overhead, making them strong base models for this task~\cite{dino,dinov2,dinov3,simclr,He2020MoCo,Wu2018InstanceDiscrimination}.\\
Biological cell imaging presents a distinct set of unsolved challenges for SSL. Unlike natural images, which exhibit easily separable class boundaries at the object level (\eg, jay vs.\ magpie~\cite{Guo2022HCSC,imagenet}), cellular microscopy images rely on subtle morphological and textural cues to distinguish similar yet distinct cell types.

%For instance, different cell types can appear indistinguishable under the same modality (e.g., B and T lymphocytes in brightfield). Conversely, the same cell type can look very different across modalities, for example, a neuron in brightfield versus the same neuron in a fluorescent ($\beta$-III-Tubulin) modality. Examples of multi-modal cell imaging are shown in Fig.~\ref{fig:hierarchy}(b). Being able to have the latent space be aware of the same cell different modality (having thsoe 2 islands closer together) and hten having better (self supervised induced) class seperation within the same modality would demonstrate a better understanding of morphology which would become useful for biologically relevant downstream tasks such as drug pertubation \& cell classification.

For instance, different cell types can appear indistinguishable under the same modality (\eg, OPCs and Neurons when captured with the same modality (Fig.~\ref{fig:hierarchy}-left)). 
%Conversely, the same cell type can look very different across modalities, for example, a neuron in brightfield versus the same neuron in a fluorescent (e.g., $\beta$-III-Tubulin) modality.
Conversely, the same cell type can appear markedly different across imaging modalities; for example, a neuron in brightfield versus the same neuron in \eg fluorescence microscopy.
A robust latent space should attempt to align the same cell type across modalities and separate different cell types within each modality, reflecting a stronger morphology awareness and supporting downstream tasks such as drug perturbation and cell classification. Examples of multi-modal cell imaging are shown in Fig.~\ref{fig:hierarchy}-left.
\par
%Yet, to build a modality-agnostic cell embedding model, these nuances need to be encoded for each provided modality.
%Yet, a modality-agnostic embedding model can only emerge if these modality-dependent nuances are explicitly encoded.
%Since the structure and state of cellular imaging is hierarchical: coarse imaging parameters precede finer biological factors such as lineage, cell type, and state.
The imaging setup, modality, and batch effects tend to be the dominant features in cell imaging\cite{haslum2024metadata,arevalo2024batch}. This encourages models to form coarse “superclusters” that obscure biologically meaningful substructure needed for tasks such as phenotype classification, mechanism-of-action (MOA) inference, or cell-state discrimination~\cite{Bendidi2024}. Supervised class-aware methods offer better performance, although they struggle with generalization due to the scarcity of large and versatile labeled datasets in cellular imaging~\cite{Huang2023}.\\
%\todo{What does lineage/hierarchical, etc. mean need example image}
%Hence,
% \begin{figure}[b]
%     \centering
%     \includesvg[width=\linewidth]{Figures/cell_imaging_radial_tree.svg}
%     \caption{Hierarchical structure of the cell-imaging corpus visualized as a radial tree, organized by imaging modality, dataset/source, cell or tissue type, and finer subtypes. \todo{carsten: fig 2 is better here. why not show a couple of examples to drive our point home? eg by showing 3 different cell types in two modalities?} \todo{carsten: where does this come from? info has to be in the caption!} }
%     \label{fig:dataset_fig}
% \end{figure}
Therefore, we propose a SSL framework that captures hierarchical structure in a single-cell representation. Our contributions are fourfold:
\begin{itemize}
    \item \textbf{Hierarchy-aware objective.}
    %We formulate a general, label-free hierarchy-aware objective that couples multi-resolution prototypes and stability-weighted anchoring from HDBSCAN~\cite{HDBSCAN} with optimized positive and negative mining, providing a simple drop-in geometric prior for SSL.
    To improve intra-supercluster decision boundaries, we propose a novel, generalisable, label-free, hierarchy-aware objective that uses an HDBSCAN cluster tree to define subcluster centroids, pulling each individual cell towards their parent clusters centroid while pushing them away from other clusters at each level of the hierarchy.

    %To improve intra-supercluster decision boundaries, we propose a novel, generalisable, label-free, hierarchy-aware objective that uses HDBSCAN-derived, stability-weighted prototypes as a simple contrastive drop-in for hierarchy learning in SSL.
    
    %We propose a novel and generalizable, label-free hierarchy-aware objective that leverages HDBSCAN-derived, stability-weighted prototypes to provide a simple contrastive drop-in framework for hierarchy learning in SSL, to improve intra-supercluster decision boundaries.
    
    %couples multi-resolution prototypes with HDBSCAN-derived~\cite{HDBSCAN} stability-weighted prototypes and optimized contrastive mining, providing a simple drop-in framework for self-supervised learning.
    \item \textbf{Double-teacher distillation.}
    To encourage morphology-aware, less modality centred representations, we add a second distillation teacher that uses segmentation masks as a weak prior to provide structure-aware supervision. It aims to pull each cell’s embedding towards its mask view to initiate breaking up modality-driven superclusters.
    
    \item \textbf{Benchmark curation.}
    %We agglomerated and processed scattered existing benchmarks into single cell, multi-modality spanning 208 different labeled classes and 2.3m cells
    We consolidated scattered benchmarks into a unified single-cell, multi-modality dataset covering 208 labeled classes and 2.3 million single cells.
    % \todo{we would need to share this via an anonymous link.}
    \item \textbf{Empirical improvements.} We outperform state-of-the-art SSL cell embedding baselines across a large multi-modality corpus, with higher retrival, clustering, and downstream metrics, reflecting better alignment with biologically meaningful subclusters.
    % \todo{VISHWA: Double check this} \footnote{Code is provided in the supplementary material.}
    %We demonstrate improvements over state-of-the-art SSL cell embedding baselines on a large-scale compilation of cell images spanning multiple modalities, reporting gains in top-$k$ accuracy and mean average precision across evaluation ranges, indicating better alignment with biologically meaningful subclusters.
\end{itemize}
Code is available at: \url{https://github.com/tum-ai/HASSL}. \\
Curated dataset is available at: \url{https://huggingface.co/datasets/tum-ai/HASSL-SingleCellBench}.

\section{Related Work}

%We review (i) general-purpose visual representation learning, (ii) microscopy- and cell-imaging–specific approaches, and (iii) hierarchical self-supervised learning (SSL). This synthesis motivates our objective: learning memory-efficient image representations that respect multiple hierarchies in unlabeled settings.
%We review three lines of work: general-purpose visual representation learning, microscopy-specific cell imaging, and hierarchical SSL, which motivate our goal of memory-efficient representations that encode multi-level structure in unlabeled data. For single-cell microscopy this means capturing both modality and fine-grained morphology under heterogeneous assays, yet standard visual self-supervision still overfits to global modality clustering and dataset bias instead of morphology-centric hierarchy.
We review visual representation learning, microscopy-specific cell imaging, and hierarchical SSL, motivating efficient representations that capture multi-level structure in unlabeled data. For single-cell microscopy, this means encoding modality and fine-grained morphology across heterogeneous assays, yet standard self-supervision often overfits to global cues rather than a morphology-centric hierarchy \cite{yao2024weakly,haslum2024metadata}.

%General-purpose visual representation learning has evolved from contrastive objectives to distillation and masked-image modeling. Contrastive methods such as SimCLR~\cite{simclr} and triplet loss~\cite{FaceNet} avoid the costly pixel-space generation step of VAEs and VQ-VAEs~\cite{kingma2013auto, van2017neural}, but instead optimize anchor embeddings toward positives and away from negatives, which in turn demands numerous comparisons and strong data augmentation~\cite{He2020MoCo, simclr}. Methods like BYOL \cite{byol} and the DINO family~\cite{dino, dinov2, dinov3} tackle this by eliminating the need for negative samples through a self-distillation framework. However, this line of work is tuned for generic image semantics and coarse class separation on datasets like ImageNet~\cite{imagenet}. It does not model hierarchical relationships, and in cell imaging, it largely clusters by modality rather than capturing subtle morphological differences within a modality, posing a major limitation for single-cell representation learning.

General-purpose visual representation learning has shifted from contrastive objectives to distillation and masked-image modeling. Contrastive methods such as SimCLR~\cite{simclr} and triplet loss~\cite{FaceNet} avoid pixel-space generation in VAEs and VQ-VAEs~\cite{kingma2013auto,van2017neural}, but require many comparisons, strong augmentations, and explicit negatives~\cite{He2020MoCo,simclr}. Distillation methods such as BYOL~\cite{byol} and the DINO family~\cite{dino,dinov2,dinov3} remove negatives via self-distillation, but are tuned for generic semantics and coarse separation on datasets like ImageNet~\cite{imagenet}. They do not model hierarchy, and in cell imaging they often cluster by modality rather than capturing subtle within-modality morphology, limiting single-cell representation learning.

Cell imaging has followed a similar shift, moving from assay-specific pipelines to generalisable SSL models. CellPaint-DINO~\cite{kim2025self} benchmarks standard SSL for Cell Painting using MAE~\cite{kraus2024masked}, DINO~\cite{dino}, and SimCLR~\cite{simclr}, with DINO as the strongest baseline. Microscopy-specific designs address practical constraints. Chada-ViT~\cite{bourriez2024chadavitchanneladaptive} handles variable channel counts via channel-aware tokenisation. SubCell~\cite{gupta2024subcell} uses MAE-style pretraining~\cite{kraus2024masked} to learn protein localisation and morphology from HPA images~\cite{hpa_single_cell_classification}. OpenPhenom~\cite{kraus2024masked} trains channel-agnostic ViT and MAE models on millions of RxRx~\cite{rxrx_openphenom} and Cell Painting images. scDINO~\cite{pfaendler2023self} adapts self-distillation to fluorescent single-cell crops on top of a DINO backbone. Yet these methods are often channel- or assay-aware rather than morphology-aware because they do not link embeddings to segmentation geometry during training, leaving morphology cues weak.

The works most similar to ours are hierarchy-aware mining methods that modify pair selection so the loss encodes structure directly CoHiClust~\cite{Znalezniak2023CoHiClust} learns a contrastive cluster tree, while SeedViews~\cite{Xu2023SeedViews} uses multi-level views with multiple positives to keep semantically similar instances aligned. HCSC~\cite{Guo2022HCSC} is closest to our approach, dynamically constructing hierarchical prototypes and adaptively expanding positives to nearby semantics while sharpening negatives for distinct samples. However, its recursive k-NN sampling relies on hard positives and negatives that are difficult to define stably and can under-emphasise fine morphology, leading to unstable gradients and slower convergence.

%Unlike previous work
Our method combines stability-weighted hierarchical prototypes with double-teacher distillation. The zero-shot segmentation masks add geometric cues that guide coarse-to-fine morphology learning, yielding hierarchy-aware embeddings that reduce modality-driven clustering.

\newcommand{\parent}{\pi}
\newcommand{\Children}{\mathrm{Ch}}

\section{Methodology}
\begin{figure}[!t]
  \centering
  \includegraphics[width=\textwidth]{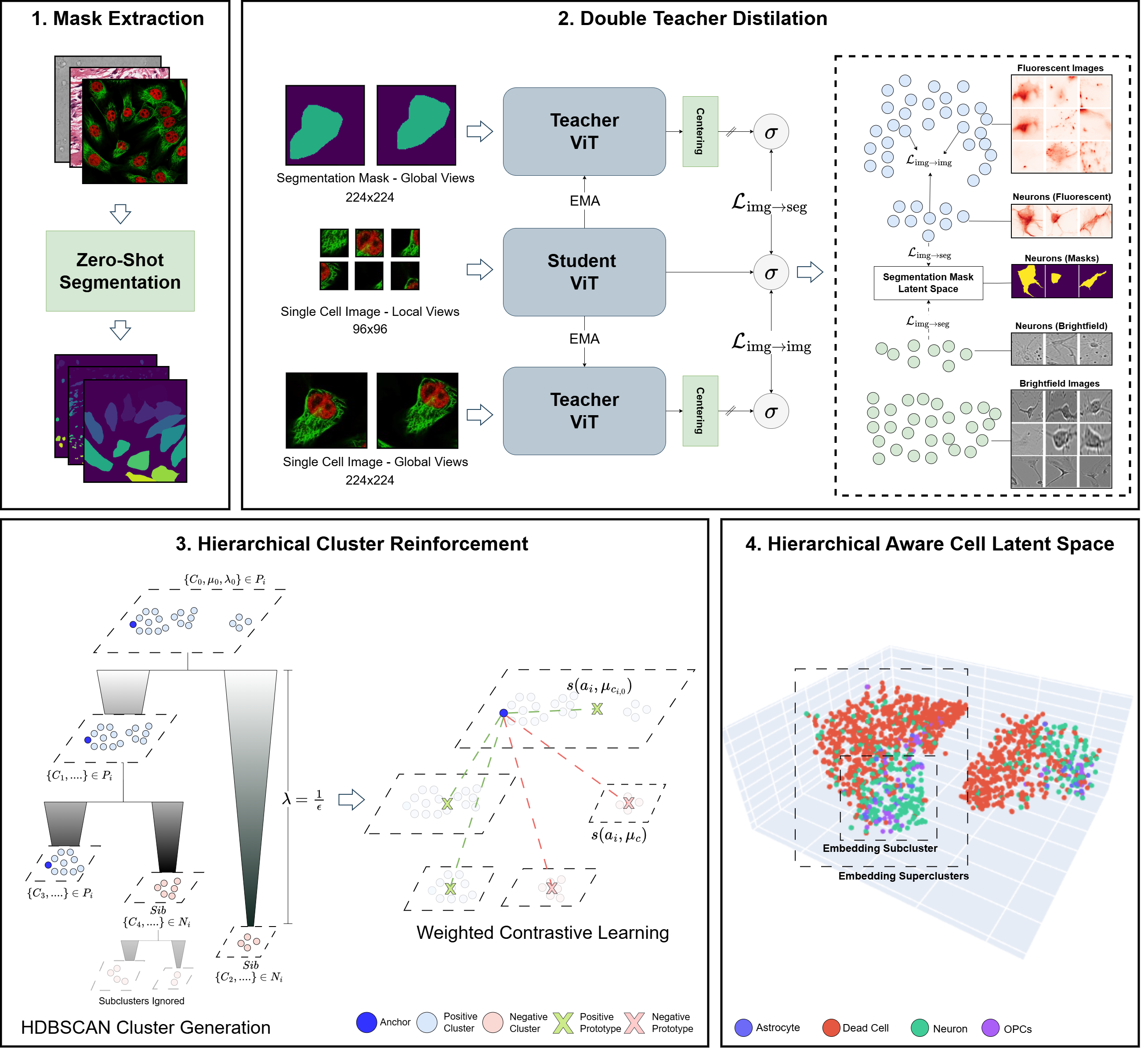}
  \caption{Overview of our hierarchy-aware representation learning pipeline.
(1) Segmentation-map generation: microscopy images are zero-shot segmented~\cite{Cellpose1}, which is used as a proxy for morphological features 
(2) Double-Teacher distillation: a student ViT is trained with EMA from a global image teacher and a segmentation teacher, promoting subcluster breakout in the latent space.
(3) Hierarchical cluster reinforcement: multi-level clusters are formed with HDBSCAN~\cite{HDBSCAN}; cluster stability 
$\lambda$ weights positive pairs within parents, while negative is defined as $\frac{1}{\lambda}$, yielding stability-weighted contrastive learning.
(4) Resulting latent space: the learned embedding organizes cells into superclusters with clearer subclusters (illustrated for astrocytes, neurons, oligodendrocyte progenitor cells (OPCs), and dead cells), reflecting improved hierarchy awareness and structure-preserving separation.}
  \label{fig:archit}
\end{figure}

%\TODO{what exactly goes in to the network? e.g. single cell image represented with birghtfield / DAPI / ... fluoprescent channel, what exactly comes our from the model? e.g. channel name / cell type (e.g. neuronal) / cell subtype (e.g. neuron, astrocyte, ...)}
%Our goal is to create a framework for representation learning models that accepts a single-cell as a single-modality RGB image and generates a one-dimensional embedding.
%We design a representation-learning framework that takes a single-modality of a single cell image and produces a d-dimensional embedding vector. This framework would allow a model to learn a hierarchy-aware latent space without any explicit information about the said hierarchy. For this, we want to rely on an approximation of the cell structure and promote the formation of subclusters more actively.

% \todo{add: Cellular microscopy images exhibit an implicit hierarchical structure, where the visual appearance of a single cell varies across molecular markers and cell types.}

We design a representation-learning framework that embeds a single modality of a single-cell image. This framework enables the model to learn a hierarchy-aware latent space without any explicit supervision about the hierarchy itself. To achieve this, we approximate cellular morphology and actively encourage the formation of meaningful subclusters in the embedding space.
%Our method needs to achieve two goals:
%\begin{itemize}
%    \item Inform the embeddings about the cell structure, agnostic to the modality of the image, via the utilization of segmentation masks. This would initiate the formation of label-free morphology-based sub-clusters.
%    \item Make the loss pull cells in the same sub-cluster together and push different ones apart for every level in the hierarchy. 
%\end{itemize}
Our method must satisfy two objectives. First, it needs to inform the embeddings about cellular structure, independent of imaging modality, by leveraging segmentation masks as weak priors. This initiates the emergence of label-free, morphology-based subclusters. Second, the loss must pull cells belonging to the same subcluster together while pushing apart cells from different subclusters across all levels of the hierarchy.
We propose an annealed training strategy with an additional teacher head to instruct the embeddings about the structure, and a loss that uses hierarchical clustering to reinforce the subcluster structures in the latent space. Figure~\ref{fig:archit} illustrates our model.

\subsection{Double-Teacher DINO Distillation}
%(Image and Segmentation Teachers)}
\label{double_teacher_methodology}
%To bias representations away from modality cues and toward morphological structure, we guide the model’s attention to cell shape. Our single-cell extraction pipeline produces segmentation masks that encode cell geometry and act as weak, structure-aware supervision. We extend the DINO distillation framework with a segmentation teacher that injects gradient flows from mask-based supervision alongside image features. This enables morphology-focused representation learning due to the model being informed on cell shape during the training process. The additional segmentation teacher supervises the student with shape-gradient targets computed from a generalist segmenter (Cellpose-SAM \cite{Cellpose1}), which has the beneficial side effect of initiating the breaking apart of modality superclusters into morphology based ones. \par 

%enter only through the dataloader,

To bias representations away from modality cues and toward morphology, we guide the model's attention towards cell shape. To avoid leaking labels, we pre-compute segmentation masks using a generalist segmentation model. These mask-derived shape cues enter only through the dataloader as weak, structure-aware supervision. As a weak prior, these masks can tolerate an acceptable level of noise and error, while still providing meaningful gradients to the model. Therefore, any zero-shot segmentation method would suffice in this case. We have used CellposeSAM \cite{Cellpose1} in our approach. With this mask, we implement a double teacher distillation framework with an additional segmentation teacher that conditions on both images and masks, supervising the student with mask-informed targets. This encourages morphology-focused representations and begins to break modality-driven superclusters into morphology-based ones. 
% \todo{mention some other seg. models that we could use to show that our model doesn't rely a lot on good seg. performance.}

As shown in Fig.~\ref{fig:archit} (Sec.~2), instead of the traditional single-teacher EMA setup in DINO (global teacher targets vs.\ cropped student views), we use two self-supervised teachers: an EMA image teacher that provides global semantic targets and a segmentation teacher that provides segmentation-aware targets. Both loss objectives \(\mathcal{L}\) aim to minimize the distance between the student ViT’s \cite{dosovitskiy2020image} local-view embedding and the segmentation and image global-view embeddings of the teacher ViTs. This design encourages cells of the same class but from different modalities to align with their segmentation-based embeddings, while allowing them to separate slightly from their broader modality-specific superclusters.

For the teacher, we compute sharpened, centered targets using Sinkhorn-Knopp \cite{SWAV}, following the approach used in DINOv3 \cite{dinov3}. This gives us our two targets $q_{img}$ and $q_{seg}$ over the global view $g$, defined for image and the segmentation map as:
\begin{equation}
q_{\text{img}}^{(g)}=\mathrm{SK}\!\big(z_{T,\text{img}}^{(g)};Temp\big),\qquad
q_{\text{seg}}^{(g)}=\mathrm{SK}\!\big(z_{T,\text{seg}}^{(g)};Temp\big),
\end{equation}

% , the global/local pair weights as defined as: % follow:
% \[
% w_g=\frac{G(G-1)}{G(G-1)+GL} \qquad
% w_l=\frac{GL}{G(G-1)+GL}
% \]

% \todo{define well: what are q and p? and all other defined variables?}
% \todo{how about showing it with $V_s$ for the student? -- Student consumes all crops, so Vt is shown as a subset of V. Its written that V includes both global and local crop} 
where $z_T$ are the teacher embeddings, and $Temp$ is the temperature for Sinkhorn-Knopp. To write all view-to-view distillation terms compactly, we functionalize the objective of DINOv3~\cite{dinov3}. For this, let $V$ be the set of all student views (global and local) and $V_t\subset V$ the teacher views (the global crops). With $CE(q,p)=-\sum_k q_k\log p_k$, the teacher targets $q^{(v_t)}$ for $v_t\in V_t$, and the student probabilities $p^{(v_s)}$ for $v_s\in V$. The image-level objective is the pooled average over all valid teacher$\to$student pairs:
\begin{equation}
\label{eq:phi_pooled}
\Phi\big(q,\{p\}\big)
\;=\;
\frac{1}{\,|V_t|\,(|V|-1)\,}
\sum_{v_t\in V_t}\;
\sum_{\substack{v_s\in V\\ v_s\neq v_t}}
CE\!\big(q^{(v_t)},\,p^{(v_s)}\big).
\end{equation}
\\

\noindent With this notation, the standard image-to-image loss term of DINO is defined as:
\begin{equation}
\label{eq:dino-functional}
\mathcal{L}_{\mathrm{img}\to\mathrm{img}}
=\Phi\!\big(q_{\mathrm{img}},\,\{p_{S,\mathrm{img}}^{(v)}\}_{v\in V}\big).
\end{equation}
\\
%The segmentation of teacher contributions is:
The teacher’s contribution to the segmentation branch is:
\begin{equation}
\label{eq:seg-to-seg}
\mathcal{L}_{\mathrm{seg}\to\mathrm{seg}}
=\Phi\!\big(q_{\mathrm{seg}},\,\{p_{S,\mathrm{seg}}^{(v)}\}_{v\in V}\big),
\end{equation}
\begin{equation}
\label{eq:img-to-seg}
\mathcal{L}_{\mathrm{img}\to\mathrm{seg}}
=\Phi\!\big(q_{\mathrm{seg}},\,\{p_{S,\mathrm{img}}^{(v)}\}_{v\in V}\big).
\end{equation}
\\
The $\mathcal{L}_{\mathrm{seg}\to\mathrm{seg}}$ stabilizes the segmentation embedding space. Otherwise, it would pull the image latent toward unmodeled/noisy segmentation targets. We combine Eqs. \ref{eq:dino-functional}, \ref{eq:seg-to-seg}, and \ref{eq:img-to-seg} via a convex combination to get our Double-Teacher objective as follows:
\begin{equation}
\label{eq:doubleteacher}
\mathcal{L}_{\text{DoubleTeacher}}
= (1 - \gamma)\,\mathcal{L}_{\mathrm{img}\to\mathrm{img}}
+ \gamma\left(\mathcal{L}_{\mathrm{seg}\to\mathrm{seg}}
+\mathcal{L}_{\mathrm{img}\to\mathrm{seg}}\right),
\end{equation}
where $\gamma\in[0,1]$.

% \todo{define all variables}

% \todo{Start with input what is output overall goal from beginning of methodlology}
\subsection{Hierarchy Aware Contrastive Loss}
%by Proxy via HDBSCAN}
\label{hdbscan_methodology_section}
%The core idea of this part of the framework is
%We want to encourage a stronger hierarchical sub-cluster structure within our latent space without the use of labels.
We aim to strengthen the hierarchical subcluster structure in the latent space without using any labels.
%In a nutshell, given the batch embeddings, we run HDBSCAN \cite{HDBSCAN} to obtain a minimum spanning tree showing, for each point, its cluster memberships from leaf to root for each level in the hierarchy.
In a nutshell, given the batch embeddings, we run HDBSCAN \cite{HDBSCAN} to obtain a minimum spanning tree (MST) that reveals each point’s cluster memberships from leaf to root across the hierarchy.
At each level of the tree, we compute a cluster centroid (“prototype”) and, for each point, mine positive and negative prototypes within our stability-$\lambda$-weighted hinge-contrastive loss. Together, the losses draw each point toward its ancestor prototypes while repelling negatives at each hierarchical level. Negatives that share the point’s parent are still drawn to that parent, but in a manner that maximizes their separation from the point while preserving latent-space integrity, resulting in clear separation between traditionally similar yet morphologically distinct cells (\eg, OPCs and Neurons in brightfield as seen in Fig.~\ref{fig:hierarchy}-left).

%\todo{On explaining hierarchy, how about mentioning: human cells and mouse cells, marker types on the cells, and their type? It would show 3 hierarchical layers.}

%In a nutshell, given DINOv3 batch embeddings, we run HDBSCAN \cite{HDBSCAN} to obtain an Minimum Spanning Tree (MST) of the latent space and compute a centroids for each cluster as targets, called prototypes. For each point, we use its parents in the MST as the positive, select negatives as defined in Sec. 3.2.2, and weight the hinge-contrastive loss by the stability $\lambda$ for each cluster.

% \todo{Make clear what is our contribution and what is default DINO. Increase intuition regarding 2nd teacher, cos rn not clear that we essentially add 2nd teacher. define terms before hand before shoving in. explain what are crops. define teacher. define shit. function form. double definition L dino, Limgimg, define domain input and output of the function, convex combination over convex addition. cross entropy.}

%In a nutshell, given DINOv3 batch embeddings, we construct a cluster-based minimum spanning tree (MST) via HDBSCAN \cite{HDBSCAN}, compute centroids as our proxy for each cluster, mine positives (MST-parents) and negatives (to be explained in section 3.2.2) for each point, and utilize our $\lambda$ stability weighting during the hinge-based contrastive loss. 
\noindent
\textbf{Multi-resolution hierarchy aware prototypes.}
Flat clustering methods like DBSCAN \cite{Actual-DBSCAN} force one resolution. They either merge subtypes or over-fragment into microclusters. Hierarchical clustering methods avoid this by keeping fine clusters while grouping them under broader parent clusters. Hence, we use \textsc{HDBSCAN} on in-memory batch embeddings to obtain the condensed cluster tree, an minimum-spanning-tree (MST) view of the latent-space hierarchy, setting \texttt{min\_cluster\_size}=2 to maximize depth while avoiding singleton leaves. This depth-first, label-free approach is needed because the categories and subcategories are unknown. Hence, the model discriminates on minute morphological differences at the leaf level, which supports subtle, biologically relevant downstream tasks. As illustrated in Fig.~\ref{fig:archit} (Sec.~3), each anchor is assigned, at every level of the hierarchy, to a cluster together with its member points, forming a nested hierarchy of memberships. Clusters that the anchor does not belong to serve as negatives. We denote the $D$ dimensional, L2-normalized student embedding of sample $i$ by $\hat{x}_i \in \mathbb{R}^D$ and define the anchor as $a_i = \hat{x}_i$. Each node $c$ in the tree corresponds to a cluster with member set $\mathcal{C}_c$ and stability (persistence) proxy $\lambda_c$. These level-wise prototypes and stabilities define the positive and negative sets used for prototype construction and mining in our contrastive objective.
%based on subtle morphological differences. To do so let $\hat{x}_i\in\mathbb{R}^D$ denote the L2-normalized student embedding (anchor $a_i=\hat{x}_i$), and each node $c$ a cluster with members $\mathcal{C}_c$ and stability proxy $\lambda_c$.

Rather than traditional instance pairs, we utilize prototypical objectives. For each positive and negative cluster in Fig.~\ref{fig:archit} (Sec.~3), we compute centroid prototypes $\mu_{c_i}$ (green and red crosses) relative to the anchor, which summarize the clusters the anchor belongs to and those it does not, and thus define our positive and negative prototypes. This lowers gradient variance and provides consistent coarse-to-fine targets, stabilizing optimization and encouraging hierarchical structure \cite{Li2021PCL,SWAV,Guo2022HCSC}. For any cluster $c$, we define the L2 normalized centroid
\begin{equation}
\mu_c \;=\; \mathrm{norm}\!\left(\frac{1}{|\mathcal{C}_c|}\sum_{j\in\mathcal{C}_c}\hat{x}_j\right)\in\mathbb{S}^{D-1}.
\end{equation}

% \todo{what are each variable? -- already defined in the preious para, and S (D-1) is a common unit sphere for L2 norm.}

% reduces contrastive mining from $\mathcal{O}(B^2)$ to $\mathcal{O}(BK)$ per batch (with $K!\ll!B$)\todo{what is K, define it},
% \todo{what will be the leaf? e.g. cell type?}
% \todo{what will be the near root? what is root?}
\subsubsection{Prototype mining.}
\label{prototype_mining}
In our hierarchical task (refer to Fig.~\ref{fig:archit} (Sec. 3) for the visualization of the process), not every centroid provides useful information. For optimizations across memory and to not undermine the initial benefits of self-distillation learning methods like DINO~\cite{dino}, we decided to mine for the minimum efficacious amount of positive and negative prototypes for our contrastive objective.
%A visualization of to be explained mining process and the tree can be found at part 3 of Fig.~\ref{fig:archit} \todo{needs rewriting / don't mention to be explained}.
For positives, we use the prototype of each parent along the leaf-to-root path of the selected node. For negatives, given our MST and computed centroids, we remove prototypes whose information is already encapsulated by others further up the hierarchy. As shown for parent cluster \(C_4\), the information of \(C_4\)'s child sub-clusters is already captured at the parent level. Non-essential child clusters are therefore ignored.
This allows us to select the minimum efficacious amount of negatives by removing possibly repetitive data points and therefore optimizing our memory footprint. Here is the mathematical formalization:

For each non-noise point $i$ marked by HDBSCAN, let $\mathcal{P}(i)=\{c_{i0},c_{i1},\dots,c_{iK}\}$ be its ancestor path from fine subcluster to coarse supercluster.
The \emph{positive anchors} are the centroids along this path,
\begin{equation}
P_i \;=\; \{\mu_{c_{ik}}: c_{ik}\in\mathcal{P}(i)\}.
\end{equation}

\noindent For negatives, we use only extracted prototypes from clusters that directly split from the anchor at each level. These are the other direct children of the parent cluster of our anchor. We define these as the sibling set of a node $c$:
\begin{equation}
\mathrm{Sib}(c)\;=\;\Children\big(\parent(c)\big)\setminus\{c\},
\end{equation}
Where $\parent(c)$ denotes the parent of $c$, and $\Children(.)$ as its children. With this, we collect the siblings encountered along the path as follows:

\begin{equation}
\mathcal{S}(i)\;=\;\bigcup_{k=0}^{K-1} \mathrm{Sib}(c_{ik})
\;=\;\bigcup_{k=0}^{K-1} \Big( \Children(c_{i,k+1})\setminus\{c_{ik}\} \Big).
\end{equation}

\noindent The negative anchors are then exactly the centroids of these sibling clusters:
\begin{equation}
N_i \;=\; \{\mu_{c} : c \in \mathcal{S}(i)\}.
\end{equation}
% and the \emph{negative anchors} are centroids from clusters not on the path,
% \[
% N_i \;=\; \{\mu_c : c\notin \mathcal{P}(i)\}.
% \]
\noindent
\textbf{Stability weighting via $\lambda$.}
% \todo{The clamp $\varepsilon$ prevents vanishing contributions from less stable (but potentially informative) leaves.}
Traditional prototype-contrastive methods assume labels and assign uniform weights to positives and negatives \cite{khosla2020supervised}. In a label-free setting, this is ill-posed because a point’s cluster-membership confidence increases toward deeper nodes in the MST; we quantify this confidence using HDBSCAN stability (persistence) $\lambda$ from the condensed tree, evaluated across density scales. With uniform weights, an instance is pushed equally toward both reliable and unreliable positives, which distorts the latent space. Likewise, treating closely related sibling subtypes as strong negatives on par with completely unrelated cell types breaks the intended hierarchy. Hence, we apply stability weighting using the $\lambda$ of HDBSCAN to preserve the natural hierarchy while maximizing separation across parents.\par

Let $\lambda_{ik}\!\equiv\!\lambda_{c_{ik}}$.
We map stabilities to normalized, clamped weights via
a generic transform $\phi_\varepsilon$:
\begin{equation}
\phi_\varepsilon(z_{ij})
=
\frac{\max\!\bigl(\varepsilon, z_{ij}\bigr)}
     {\sum_{u} \max\!\bigl(\varepsilon, z_{iu}\bigr)},
\end{equation}
with small $\varepsilon>0$ for numerical stability. Keeping in mind that we invert the lambdas for the negative samples, we define the positive weights ($\alpha_{ik}$) and negative weights ($\beta_{ic}$) as follows:
\begin{equation}
\alpha_{ik}
=
\phi_\varepsilon\!\left(
\frac{\lambda_{ik}-\lambda_{\min}}{\lambda_{\max}-\lambda_{\min}}
\right), 
\beta_{ic}
=
\phi_\varepsilon\!\left(
\frac{\lambda_{\max}-\lambda_{ic}}{\lambda_{\max}-\lambda_{\min}}
\right).
\end{equation}
% For negatives, we invert the stability so that more distant
% (low-stability) prototypes receive higher weight:
% \[

% \]

% \todo{define all the variables -- all variables have been defined perviously in the paragraphs. Redefining them will increase the content length and we are already half a page over}

\noindent With cosine similarity $s(u,v)=u^\top v$ and margin $m>0$, we aggregate positives and negatives by a stability-weighted mean:

\begin{equation}
s_{\mathrm{ap}}(i)=\sum_{k=0}^{K}\alpha_{ik}\,s(a_i,\mu_{c_{ik}}),
\:
s_{\mathrm{an}}(i)
= \sum_{c\in \mathcal{S}(i)} \beta_{ic}\, s(a_i,\mu_{c}),
\end{equation}

\noindent and define the per-anchor hinge loss as:

\begin{equation}
\begin{split}
L_i \;=\; \big[m + s_{\mathrm{an}}(i) - s_{\mathrm{ap}}(i)\big]_+, \quad
\mathcal{L}_{\mathrm{HDBSCAN}}=\frac{1}{|\mathcal{I}|}\sum_{i\in\mathcal{I}} L_i,
\end{split}
\label{eq:hdbloss}
\end{equation}

\noindent where $\mathcal{I}$ indexes anchors with at least one positive and one negative.

In the objective in Eq.~\ref{eq:hdbloss}, maximizing $s_{\mathrm{ap}}$ pulls $a_i$ toward a stability-weighted barycenter of its sub-/supercluster anchors, tightening intra-subcluster spread while preserving alignment with the enclosing supercluster.
Minimizing $s_{\mathrm{an}}$ repels $a_i$ from the competing sibling cluster.
Together, the hinge margin enforces
\begin{equation}
s(a_i,\text{own path}) \;\ge\; s(a_i,\text{sibling}) + m,
\end{equation}
which (i) compresses subclusters, (ii) preserves supercluster structure, and (iii) widens gaps at ambiguous boundaries.

Noise points (unclustered by HDBSCAN) are excluded from $\mathcal{I}$ and $\lambda_{\min},\lambda_{\max}$ are computed per batch.%, per path, or over a running window; the choice controls how \emph{relative} the stability scale is.
%A temperature/exponent $\chi$ can be added to $w_{ik}$ (replacing $w_{ik}$ with $w_{ik}^\chi$ before normalization) to anneal emphasis from coarse ($\chi\!\ll\!1$) to fine ($\chi\!>\!1$) anchors over training.
%Stability weighting via lambda:
%* Since we are in a unsuprevised task treating eveyrthing as a hard positive / hard negative does not make any sense. 

% which architectures like BYOL and DINO have done extremely well by utilizing self-distillation instead of contrastive learning.
    
%    (something which BYOL)
    
%Narrative:
%* For each batch the embeddings are already precomputed by DINO
%* Hence no need to recompute (easier in terms of compute)
%* We get hierarchical structure 
% - down to the level of min cluster size = 2
% * We mine positives and negatives for each point trying to minimize unessary ones for memory efficiency
% * We mine stability lambdas
% * We use the basic contrastive loss to pull together and move away
\section{Experiments and Results}

\subsection{Dataset}
\label{sec:dataset} 

Popular cell-biology benchmarks (e.g., MedMNIST/PathMNIST)~\cite{Yang2023MedMNISTv2} are not truly single-cell: images contain multiple instances and lack per-cell IDs/masks, preventing instance-level embeddings. We therefore convert the instance masks to oriented boxes and crop one cell per image, turning multi-cell datasets into a unified single-cell crop suite. Labeled sources include ~\cite{graham2021coniccolonnucleiidentification,Vu2019Methods,Naylor2019DistanceMap,MAHBOD2021104349,wang2024simultaneously,Pfaendler2022_iPSC_Morpho,naji_hussein_2023_8065174,9446924} \cite{mahbod2023nuinsseg,Cutler2022,gamper2019pannuke,gamper2020pannuke,Severin2021_PBMC_Morphology,sartorius-cell-instance-segmentation}. Unlabeled sources include~\cite{DEPTO2021101653,Cellpose1,CellposeSAM,data-science-bowl-2018,DynamicNuclearNetSegmentation_v1_0,DeepCell_TissueNet_1_1,8880654,NeurIPS-CellSeg,Dietler2020} and use the dataset name as a pseudo-label for retrieval. The resulting collection spans both deep (depth$>1$) and flat (depth$=1$) hierarchies, promoting robust generalization. \\

After dedup-free aggregation, we obtain 2,390,832 single-cell crops from 20 instance-segmentation benchmarks,  spanning eight imaging modalities with overlapping cell types. For a detailed breakdown of the modalities and classes, refer to the supplementary material. We split 90/10 within each dataset and pool the 10\% into a single test set where all datasets are represented.

\subsection{Training Methodology and Ablations}

\noindent{\textbf{Model details.}}
The Model backbone is ViT\textsubscript{S}/16 \cite{dosovitskiy2020image} with an EMA image teacher, a segmentation-teacher loss (Sec. ~\ref{double_teacher_methodology}) whose weight $\gamma$ is linearly ramped from $0$ to $0.2$ over pre-training, and our additional HDBSCAN-based (Sec.~\ref{hdbscan_methodology_section}) whose weight is ramped from $0$ to $0.1$ in the final 20 epochs; all remaining losses follow the default DINOv3~\cite{dinov3} configuration.
%DINO~\cite{dino} and iBOT~\cite{iBOT} use separate 3-layer heads (hidden 2048, bottleneck 256, $65{,}536$ prototypes). Teacher outputs are centered with Sinkhorn–Knopp \todo{}. Pretraining losses are weighted as: DINO $1.0$, iBOT $1.0$, KoLeo \cite{Sablayrolles2019Spreading} $0.1$ (Gram disabled). For hierarchy, we use a centroid-contrastive objective over density clusters (default weight $1.0$; in the HDBSCAN fine-tune variant, we set it to $0.1$).
%\todo{make sure that we set a value for all the defined parameters.}
\par
\noindent{\textbf{Training details.}}
We follow DINOv3~\cite{dinov3} defaults, training on our training split (Sec.~\ref{sec:dataset}) for 100 epochs and linearly increasing the HDBSCAN components contribution to $\lambda$=1 for 20 (batch 128, multi-crop: 2 global + 8 local) on $2\times$ NVIDIA RTX~A6000 (48GB).\\
%We follow the DINOv3~\cite{dinov3} optimization and augmentation defaults, training on the training split of our dataset (Sec.~\ref{sec:dataset}) for 100 epochs and then fine-tuning for 20 epochs with a batch size of 128 using a standard multi-crop setup (2 global and 8 local crops). All experiments are run on $2\times$ NVIDIA RTX~A6000 (48GB) GPUs.
\noindent{\textbf{Baselines and Ablations.}}
% \todo{CLARIFY THE PURPOSE OF EACH ABLATION SO ITS VISIBLE IN THE RESULTS -- WE DONT WANT THE DBSCAN CONFUSION AGAIN}
We compare our models with several cell representations and SSL approaches. These include other cellular imaging centric DINO-based models like Cellpaint DINO \cite{kim2025self} and scDINO \cite{pfaendler2023self}, HCSC\cite{Guo2022HCSC} for an alternative hierarchical objective, and models using non-DINO frameworks, such as OpenPhenom \cite{rxrx_openphenom} and ChadaViT \cite{bourriez2024chadavitchanneladaptive}. Some of these, which have not seen cellular datasets before, have been trained on our set with the same training parameters and resources. This is to ensure fairness in comparing the models.\par
Apart from comparing with other cell representation approaches and our baseline DINOv3, we also perform ablations on certain components in our approach. To validate the need for hierarchy, we evaluate DINOv3 fine-tuned with DBSCAN. Furthermore, to see the effect of stability weighting, we also evaluate an unweighted HDBSCAN fine-tuned model. Lastly, we apply weighted HDBSCAN and Double Teacher separately to measure the contribution of these components individually. All use the same backbone, crops, and optimization as above.

\subsection{Evaluation Method}
We assess embedding quality with a k-NN retrieval protocol on the global embedding space over all datasets. For each test cell, we retrieve its top-$K$ cosine neighbors ($K\in{1,3,5,9}$), exclude the query, and compute the top-$K$ accuracy (Acc@K), precision (Prec@K), and mean average precision (mAP), treating same-type cells as positives and others as negatives. Average Precision (AP) is computed on the ranked list truncated at the largest $K$, with queries lacking a same-type neighbor assigned $\texttt{AP}=0$. This evaluation emphasizes neighborhood coherence and nearest-neighbor exactness, aligning with our goal of improving subcluster consistency across modalities. \par
Furthermore, to gauge the quality of the latent space, we also test with cluster-specific metrics. For this, we use Adjusted and Normalized Mutual Information (AMI and NMI) \cite{AMI} to identify the amount of class information captured by the latent space.
% \todo{add higher is better -- already show in the tables}

% \todo{ADD THE LABELED vs UNLABELED plot (Convert this to Deep vs Shalow hierarchy instead)}

%We evaluate the embedding quality with a k-NN retrieval protocol, applied for each embedding on the global embedding space spanning all datasets. For each test data point, we retrieve the top-$K$ nearest neighbors using cosine distance, excluding the queried data point itself, and calculate the Top-$K$ accuracy (Acc@K), precision (Prec@K), and mean average precision (mAP). Cells of the same type are treated as positives, while cells from different types are treated as negatives. AP is computed on the ranked neighbor list truncated at the largest reported $K$ (here $K\in{1,3,5,9}$); queries with no other example of the same label contribute \(\texttt{AP}=0\).
%The k-NN retrieval protocol emphasizes neighborhood coherence in addition to nearest-neighbor exactness, matching our objective of improving subcluster consistency across modalities.
%%A \todo{make more clear using global latent space not per single dataset (idiot proofing the paper)}

\subsection{Downstream Cell-Type Classification}
%\begin{figure}[b]
%    \centering
%    \includegraphics[width=\linewidth]{Figures/LvU.png}
%    \caption{Comparison of the top-K accuracy of the labeled and unlabeled data. Our method performs 1.5\% better on the labeled and 5.3\% better on the unlabeled subset.}
%    \label{fig:type_acc}
%\end{figure}
%To further assess the discriminative quality of the learned embeddings, we evaluate their performance with a lightweight multilayer perceptron (MLP). The utilized MLP consists of two hidden layers with ReLU activations and a dropout rate of 0.2, and is optimized using cross-entropy loss.
To further assess the discriminative quality and generalizability of the learned embeddings, we train and evaluate a multilayer perceptron (MLP) classifier on embeddings from frozen models. The MLP has two hidden layers, ReLU activations, a 0.2 dropout rate, and is optimized via a cross-entropy loss.

In addition to our curated dataset, we introduce an unseen evaluation set, the Human Protein Atlas (HPA)~\cite{hpa_single_cell_classification} and the Allen Institute of Cell Science Perturbation Set \cite{allen_cell_drug_perturbation_2018}, which was excluded from the training pipeline to assess the generalization capability of HASSL. All evaluations use the same MLP architecture and hyperparameters to ensure comparability across datasets. We report classification accuracy, macro F1-score, as well as weighted F1-score, for both our dataset and HPA. Since accuracy can be misleading under class imbalance, macro F1 provides a class-balanced view by averaging per-class F1-scores uniformly, while weighted F1 accounts for class prevalence. Reporting both thus captures performance on rare classes (macro) and reflects robustness on the empirical distribution (weighted).

\begin{figure}[!t]
    \centering
    \includegraphics[width=\linewidth]{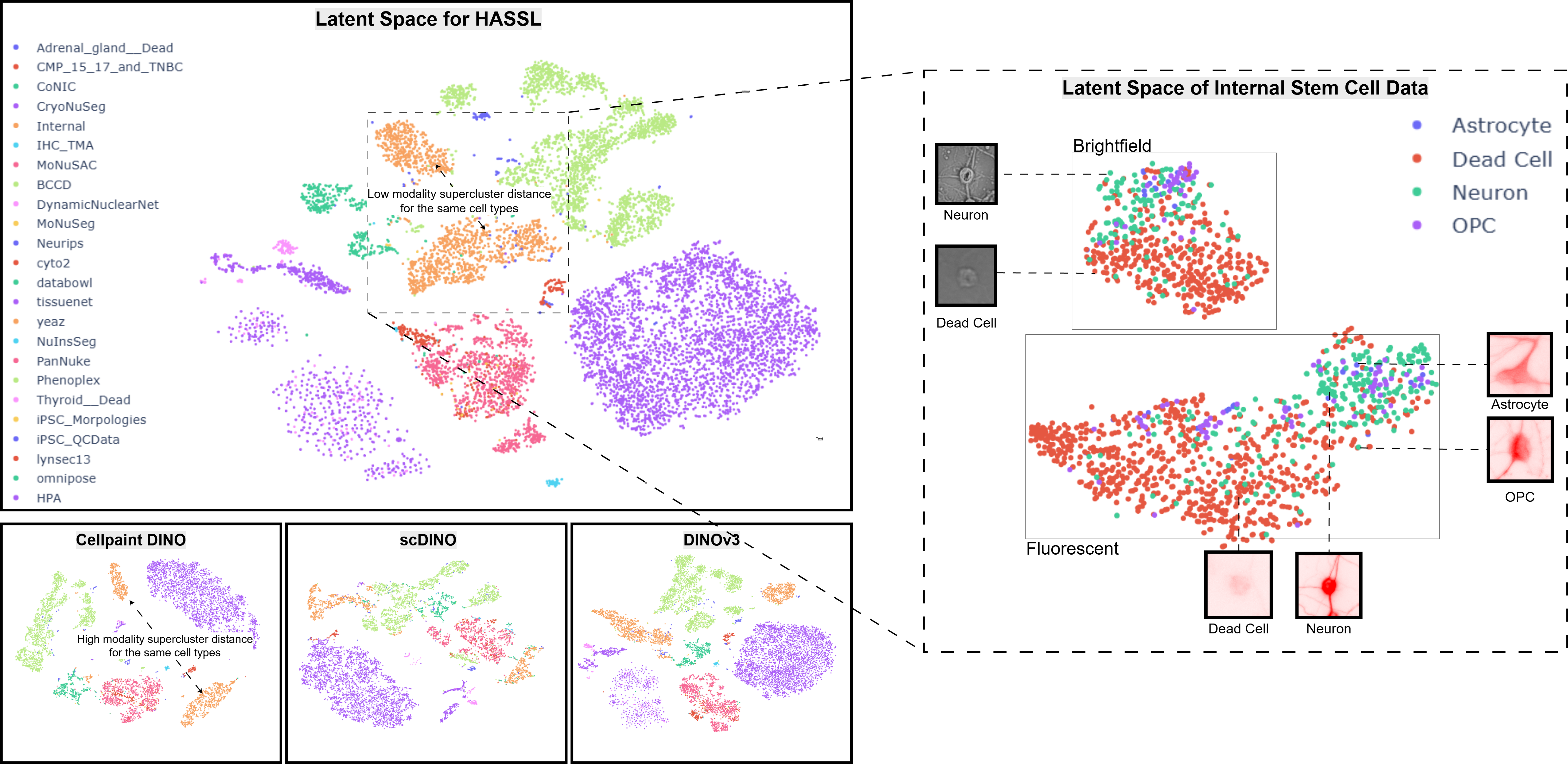}
    \caption{Latent space visualization from our model on the test set, produced using t-SNE \cite{maaten2008visualizing} with dimension=2. In the internal stem cell data (orange), the double teacher distillation brings superclusters of different modalities but the same cell type closer, and HDBSCAN still keeps the subclusters compact and discernible. In the global space, the Florucent Cluster is the 2nd closest neighbour to the Brightfield cluster for HASSL, versus 8th for DINOv3 and 22nd for scDINO.}
    \label{fig:latent_space}
\end{figure}

\subsection{Results} %explanation}

% \todo{did we define seg2img and seg2seg in the methods section?}.
\begin{table}[!t]
\centering
\footnotesize
\setlength{\tabcolsep}{4pt}
\resizebox{\textwidth}{!}{%
\begin{tabular}{@{}lccccccccccc@{}}
\toprule
\textbf{Method} &
\multicolumn{2}{c}{\textbf{K=1}} &
\multicolumn{2}{c}{\textbf{K=3}} &
\multicolumn{2}{c}{\textbf{K=5}} &
\multicolumn{2}{c}{\textbf{K=9}} &
\textbf{mAP} &
\textbf{NMI $\uparrow$} &
\textbf{AMI $\uparrow$} \\
\cmidrule(lr){2-3}\cmidrule(lr){4-5}\cmidrule(lr){6-7}\cmidrule(lr){8-9}
& Acc & Prec & Acc & Prec & Acc & Prec & Acc & Prec & & & \\
\midrule
Cellpaint-DINO~\cite{kim2025self} &
44.8 & 44.8 & 55.7 & 43.8 & 61.7 & 43.2 & 68.9 & 42.4 & 49.7 & 47.0 & 46.4 \\
ChadaViT~\cite{bourriez2024chadavitchanneladaptive} &
46.3 & 46.3 & 58.1 & 44.8 & 71.7 & 45.4 & 79.4 & 43.8 & 52.9 & 45.2 & 44.6 \\
OpenPhenom~\cite{rxrx_openphenom} &
45.3 & 45.3 & 57.1 & 35.1 & 63.9 & 32.6 & 71.9 & 30.5 & 50.8 & 40.3 & 39.5 \\
scDINO~\cite{pfaendler2023self} &
50.4 & 50.4 & 61.9 & 49.2 & 68.0 & 48.5 & 75.3 & 47.5 & 55.4 & 43.9 & 43.3 \\
HCSC~\cite{Guo2022HCSC}* &
46.1 & 46.1 & 61.6 & 45.3 & 64.5 & 44.7 & 79.7 & 40.7 & 53.9 & 45.3 & 44.6 \\
\midrule
Baseline DINOv3~\cite{dinov3}* &
49.4 & 49.4 & 64.4 & 48.5 & 72.1 & 48.0 & 80.9 & 47.4 & 56.5 & 46.8 & 46.2 \\
DINOv3 + DBSCAN &
49.1 & 49.1 & 61.6 & 47.6 & 68.1 & 46.7 & 75.6 & 45.2 & 55.2 & 47.2 & 46.6 \\
DINOv3 + Unweighted HDBSCAN &
48.4 & 48.4 & 60.6 & 46.7 & 67.2 & 45.8 & 74.5 & 44.7 & 54.3 & 47.0 & 46.3 \\
\midrule
HASSL (without Double Teacher) &
\textbf{50.9} & \textbf{50.9} & 66.6 & \textbf{50.1} & 74.3 & \textbf{49.5} & 82.8 & \textbf{48.7} & \textbf{58.1} & 47.6 & 46.9 \\
HASSL (without HDBSCAN) &
50.6 & 50.6 & 67.6 & 49.5 & 75.1 & 48.9 & 82.9 & 48.0 & 57.4 & 47.2 & 46.2 \\
\textbf{HASSL} &
\textbf{50.9} & \textbf{50.9} & \textbf{68.0} & 50.0 & \textbf{75.5} & 49.4 & \textbf{83.5} & 48.6 & 57.8 & \textbf{47.9} & \textbf{47.3} \\
\bottomrule
\end{tabular}%
}
\caption{Top-$k$ retrieval and clustering agreement results (\%). We report Acc/Prec for $K\in\{1,3,5,9\}$, overall mAP, and clustering agreement metrics (NMI and AMI). Relative to the DINOv3 baseline, stability-weighted HDBSCAN and Double-Teacher distillation provide complementary gains in retrieval performance, and their combination yields the strongest clustering agreement. \textit{* retrained on our subset for fairness}.}
\label{tab:overall_topk}
\end{table}

\subsubsection{Retrieval and Latent Space.}
\begin{figure}[!b]
    \centering
    \includegraphics[width=\linewidth]{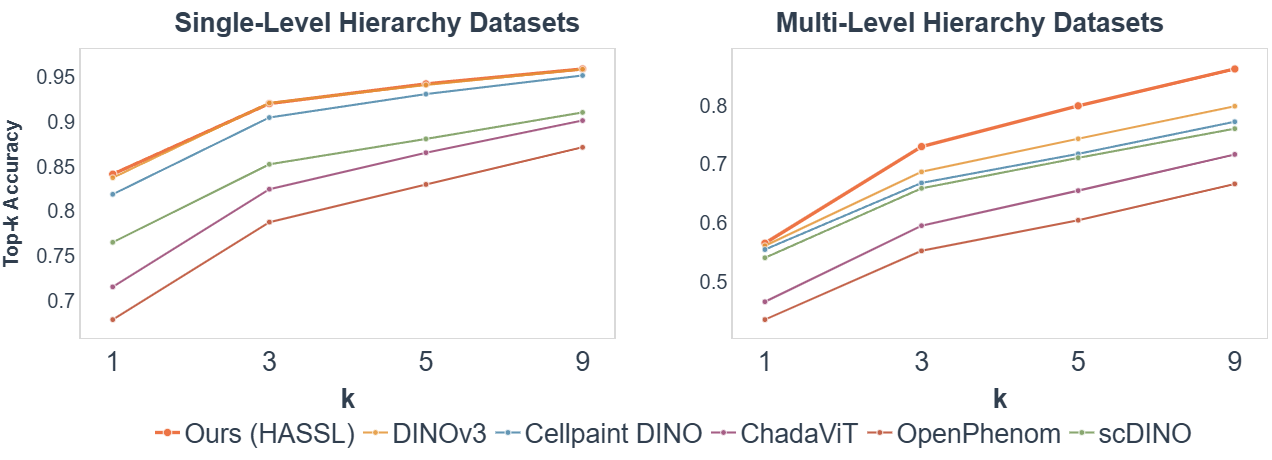}
    \caption{Top-k retrieval accuracies for subsets containing single-level hierarchy datasets (left, depth=1) and multi-level hierarchy datasets (right, depth>1). Here, HASSL is the best performing model, with an improvement of 6.3\% on multi-level hierarchy datasets.}
    \label{fig:hie-topk}
\end{figure}
%Table~\ref{tab:overall_topk} benchmarks our variants against state-of-the-art self-supervised models across accuracy@K, precision@K, and mAP. Across all variants, the two proposed components, Double-Teacher distillation and Weighted HDBSCAN, consistently improve the structure of the learned embedding space. The segmentation teacher primarily enhances mid-range neighborhood quality (e.g.\ $K=3$–$5$), reducing modality-driven drift and aligning cells with similar morphology even when their imaging appearance differs. In contrast, the Weighted HDBSCAN loss produces clearer and more compact subclusters, improving large-$K$ retrieval by enforcing coherence within hierarchical groupings. These behaviours match the design intention, which was to encourage morphology-aligned embeddings through segmentation maps, while stability weighing prototypes to refine the fine hierarchical layout.

%Replacing HDBSCAN with a flat DBSCAN clustering significantly weakens the effect \todo{what effect?}. Without hierarchical information, the model cannot distinguish between fine and coarse-grained substructures, which leads to less compact clusters and smaller gains in retrieval. This demonstrates that hierarchical supervision is essential for shaping multi-resolution structure in the embedding space.

Table~\ref{tab:overall_topk} reports accuracy@K, precision@K, mAP, and the clustering metrics (NMI and AMI) for our variants and self-supervised baselines. Both components, double-teacher distillation and weighted HDBSCAN, consistently boost retrieval: the segmentation teacher reduces the modality-driven separation and improves top-range neighborhoods ($K{=}5$–$9$), while stability-weighted HDBSCAN sharpens hierarchical subclusters and improves small to mid-$K$ retrieval. Ablations with flat DBSCAN or unweighted HDBSCAN largely remove these gains, showing that treating all hierarchy levels equally weakens boundaries and multi-resolution structure. Together, the two components are complementary and yield consistent improvements across all $K$ and a latent space that preserves modality superclusters but exhibits crisper type-level groupings (Fig.~\ref{fig:latent_space}). To further show the effect of this approach on deep hierarchies, we divide our datasets into single-level hierarchy (depth=1) and multi-level hierarchy (depth>1) subsets, and show their retrieval accuracy in Fig.~\ref{fig:hie-topk}. The plots show that HASSL improves considerably over the baseline ($6.3\%$ for $k=9$) without compromising its performance on datasets with flat hierarchies. The combination also yields the highest clustering agreement across all metrics.

\noindent
\textbf{Downstream cell classification on our curated dataset.}
%%A \todo{@Ani can you summarize this thing? I think two paragraphs is too much to explain this.
The purpose of this test is to show the model's ability to recognize cell morphologies amongst the modalities and morphologies it has seen. As shown in Table ~\ref{tab:nn_classifier_results}, HASSL yields the strongest separability on frozen features. Macro F1 improves because the hierarchy-aware loss reduces confusion among small subclasses, while weighted F1 remains competitive as coarse cluster structure is preserved.

% \medskip
\begin{table}[!t]
\centering
\scriptsize
\setlength{\tabcolsep}{2pt}
\renewcommand{\arraystretch}{0.95}

\begin{adjustbox}{max width=\linewidth,center}
\begin{tabular}{@{}lccc@{}}
\toprule
\textbf{Method} & \textbf{Acc (\%)} & \textbf{F1$_\text{macro}$ (\%)} & \textbf{F1$_\text{weighted}$ (\%)} \\
\midrule

\textbf{HASSL (Ours)} &
$\mathbf{45.6}{\tiny\,(0.0)}$ &
$48.8{\tiny\,(0.3)}$ &
$\mathbf{43.6}{\tiny\,(0.0)}$ \\

Baseline DINOv3~\cite{dinov3} &
$44.9{\tiny\,(0.1)}$ &
$47.6{\tiny\,(0.2)}$ &
$42.8{\tiny\,(0.1)}$ \\

HASSL (w/o DT) &
$44.8{\tiny\,(0.1)}$ &
$46.5{\tiny\,(0.1)}$ &
$42.9{\tiny\,(0.2)}$ \\

HASSL (w/o HDBSCAN) &
$44.6{\tiny\,(0.1)}$ &
$47.8{\tiny\,(0.0)}$ &
$42.6{\tiny\,(0.2)}$ \\

Cellpaint-DINO~\cite{kim2025self} &
$43.4{\tiny\,(0.3)}$ &
$\mathbf{49.4}{\tiny\,(0.2)}$ &
$41.4{\tiny\,(0.3)}$ \\

ChadaViT~\cite{bourriez2024chadavitchanneladaptive} &
$42.6{\tiny\,(0.1)}$ &
$46.1{\tiny\,(0.2)}$ &
$40.6{\tiny\,(0.1)}$ \\

HCSC~\cite{Guo2022HCSC} &
$41.9{\tiny\,(0.1)}$ &
$43.1{\tiny\,(0.1)}$ &
$41.9{\tiny\,(0.3)}$ \\

scDINO~\cite{pfaendler2023self} &
$37.6{\tiny\,(0.0)}$ &
$37.7{\tiny\,(0.9)}$ &
$34.4{\tiny\,(0.1)}$ \\

OpenPhenom~\cite{rxrx_openphenom} &
$34.7{\tiny\,(0.1)}$ &
$28.6{\tiny\,(0.1)}$ &
$31.7{\tiny\,(0.2)}$ \\
\bottomrule
\end{tabular}
\end{adjustbox}

\caption{Downstream multi-class classification on our dataset, sorted by F1$_\text{macro}$ (high $\rightarrow$ low). Values are mean (standard deviation) across 5 folds, reported as percentages. \textbf{HASSL} achieves the best accuracy and F1$_\text{weighted}$, and improves over Baseline DINOv3 by ${+}0.7$ Acc, ${+}1.2$ F1$_\text{macro}$, and ${+}0.8$ F1$_\text{weighted}$, while remaining within $0.6$ F1$_\text{macro}$ of the best Cellpaint-DINO.}
\label{tab:nn_classifier_results}
\end{table}

% Training the same architecture on a fixed train-test split (Table~\ref{tab:downstream_cls_hpa}) yields a different ranking than on the original dataset: \texttt{Cellpaint DINO} attains the highest accuracy (0.547; macro-F1 0.534). Our Weighted HDBSCAN + Double Teacher variant achieves 0.421 accuracy (macro-F1 0.399), outperforming Baseline DINOv3 (0.394; 0.373) . These outcomes suggest that HPA’s morphology and class taxonomy favor models tuned toward its distribution, while our hierarchy-aware refinement still provides consistent gains over the plain DINOv3 backbone on HPA. 
\noindent
\textbf{Drug identification from perturbed cell images.}
%On the HPA~\cite{hpa_single_cell_classification} single cell classification dataset, a model specifically tuned to HPA-like data, Cellpaint DINO~\cite{kim2025self} performs best (Tab.~\ref{tab:downstream_cls_hpa}). It was trained on JUMP-CP \cite{jump_cp}, which contains immunofluorescent dyes and genetic perturbations to organelles, resulting in significant carryover to the HPA task of organelle classification.
The main goal with this task is to measure the model's capability of understanding morphological differences in a deeply hierarchical dataset. For this, we have used immunofluorescent images from the Perturbation Dataset from the Allen Institute of Cell Science \cite{allen_cell_drug_perturbation_2018} containing 7 cell lines (AICS-10/12/16/22/23/24/25) perturbed with  2 drugs - paclitaxol and brefeldin. We frame the task such that a multi-layer perceptron takes in the embedding of the cell image and predicts the drug it was perturbed with (or if the image was a control). Compared to standard cell classification, this downstream gives us a more biologically relevant benchmark, as modeling the biochemistry behind drug perturbation is often challenging and very important for drug discovery. The results in Table~\ref{tab:downstream_cls_pert} show that HASSL demonstrates a considerable improvement over the baselines, demonstrating its capabilities in understanding hierarchy and identifying small morphological differences within the same cell lines.   

% --- Table 3: Perturbation ---
\begin{table}[!t]
\centering
\setlength{\tabcolsep}{2pt}
\renewcommand{\arraystretch}{0.95}
\scriptsize
\begin{tabular}{@{}lccc@{}}
\toprule
\textbf{Method} & \textbf{Acc (\%)} & \textbf{F1$_\text{macro}$ (\%)} & \textbf{F1$_\text{weighted}$ (\%)} \\
\midrule

\textbf{HASSL (ours)} &
$\mathbf{92.2}{\tiny\,(0.4)}$ &
$\mathbf{88.9}{\tiny\,(0.8)}$ &
$\mathbf{92.0}{\tiny\,(0.5)}$ \\

HASSL (w/o DT) &
$88.8{\tiny\,(0.3)}$ &
$86.7{\tiny\,(0.9)}$ &
$88.3{\tiny\,(0.5)}$ \\

OpenPhenom~\cite{rxrx_openphenom} &
$85.7{\tiny\,(0.4)}$ &
$75.8{\tiny\,(0.6)}$ &
$84.2{\tiny\,(0.4)}$ \\

HASSL (w/o DT) &
$83.0{\tiny\,(0.1)}$ &
$79.0{\tiny\,(0.1)}$ &
$82.8{\tiny\,(0.1)}$ \\

Baseline DINOv3~\cite{dinov3} &
$81.9{\tiny\,(0.2)}$ &
$77.1{\tiny\,(0.3)}$ &
$81.8{\tiny\,(0.2)}$ \\

Cellpaint-DINO~\cite{kim2025self} &
$84.2{\tiny\,(0.4)}$ &
$71.8{\tiny\,(1.3)}$ &
$81.7{\tiny\,(0.7)}$ \\

ChadaViT~\cite{bourriez2024chadavitchanneladaptive} &
$79.2{\tiny\,(0.5)}$ &
$72.5{\tiny\,(0.9)}$ &
$78.3{\tiny\,(0.6)}$ \\

HCSC~\cite{Guo2022HCSC} &
$64.4{\tiny\,(1.2)}$ &
$56.2{\tiny\,(2.5)}$ &
$63.5{\tiny\,(1.3)}$ \\

scDINO~\cite{pfaendler2023self} &
$65.0{\tiny\,(1.2)}$ &
$53.8{\tiny\,(2.6)}$ &
$63.4{\tiny\,(1.4)}$ \\

\bottomrule
\end{tabular}
\caption{Downstream perturbation identification task. Values represent the mean with standard deviation in parentheses across 5 folds, reported as percentages. \textbf{HASSL} improves considerably over the other baselines, beating the next best model, Cellpaint-DINO, by ${+}7.8$ in F1$_\text{weighted}$.}
\label{tab:downstream_cls_pert}
\end{table}

% % --- Table 2: Endpoint ---
% \begin{table}[t]
% \centering
% \label{tab:downstream_cls_endpoint}
% \setlength{\tabcolsep}{2pt}
% \renewcommand{\arraystretch}{0.95}
% \footnotesize
% \resizebox{\columnwidth}{!}{%
% \begin{tabular}{@{}lccccc@{}}
% \toprule
% \textbf{Method} & \textbf{Acc} & \textbf{F1$_\text{macro}$} & \textbf{Prec$_\text{macro}$} & \textbf{Rec$_\text{macro}$} & \textbf{F1$_\text{weighted}$} \\
% \midrule
% Cellpaint DINO                      & \textbf{0.496} & \textbf{0.461} & \textbf{0.460} & \textbf{0.470} & \textbf{0.483} \\
% Weighted HDBSCAN + Double Teacher   & 0.407 & 0.373 & 0.376 & 0.380 & 0.398 \\
% scDINO                              & 0.382 & 0.347 & 0.372 & 0.353 & 0.371 \\
% Baseline DINOv3                     & 0.380 & 0.344 & 0.349 & 0.354 & 0.369 \\
% OpenPhenom                          & 0.330 & 0.297 & 0.311 & 0.304 & 0.317 \\
% \bottomrule
% \end{tabular}%
% }
% \caption{Downstream multi-class classification on our stem cell dataset ($C{=}4$) with no Fine-Tuning}
% \end{table}

\noindent
\textbf{Human Protein Atlas (HPA) dataset.} Although HPA~\cite{hpa_single_cell_classification} does not show a deep hierarchy, it has a vastly different modality with the immunofluorescent images and cell types that the model has not seen. This gives us a good insight into the model's capability to generalize to a different type of data. Table ~\ref{tab:downstream_cls_hpa} shows that the model tuned to HPA-like data, CellPaint-DINO~\cite{kim2025self}, achieves the best performance. This is expected: it was trained on the massive JUMP-CP~\cite{jump_cp} dataset, whose immunofluorescent dyes and organelle-targeted genetic perturbations closely match the imaging characteristics of HPA. As a result, CellPaint-DINO has effectively “seen” similar organelle structures during training, giving it a clear advantage on this task.

\begin{table}[!t]
\centering
\setlength{\tabcolsep}{2pt}
\renewcommand{\arraystretch}{0.95}
\scriptsize
\begin{tabular}{@{}lccc@{}}
\toprule
\textbf{Method} & \textbf{Acc (\%)} & \textbf{F1$_\text{macro}$ (\%)} & \textbf{F1$_\text{weighted}$ (\%)} \\
\midrule

Cellpaint-DINO~\cite{kim2025self} &
$\mathbf{54.8}{\tiny\,(0.2)}$ &
$\mathbf{53.5}{\tiny\,(0.3)}$ &
$\mathbf{54.8}{\tiny\,(0.3)}$ \\

\textbf{HASSL (ours)} &
$54.2{\tiny\,(0.3)}$ &
$52.7{\tiny\,(0.4)}$ &
$54.2{\tiny\,(0.4)}$ \\

HASSL (w/o HDBSCAN) &
$53.9{\tiny\,(0.3)}$ &
$52.2{\tiny\,(0.5)}$ &
$53.9{\tiny\,(0.4)}$ \\

Baseline DINOv3~\cite{dinov3} &
$51.5{\tiny\,(0.3)}$ &
$50.0{\tiny\,(0.3)}$ &
$51.5{\tiny\,(0.3)}$ \\

HASSL (w/o DT) &
$50.1{\tiny\,(0.3)}$ &
$48.4{\tiny\,(0.4)}$ &
$50.0{\tiny\,(0.4)}$ \\

HCSC~\cite{Guo2022HCSC} &
$39.6{\tiny\,(0.3)}$ &
$38.0{\tiny\,(0.4)}$ &
$39.5{\tiny\,(0.3)}$ \\

scDINO~\cite{pfaendler2023self} &
$39.5{\tiny\,(0.1)}$ &
$37.4{\tiny\,(0.3)}$ &
$39.3{\tiny\,(0.3)}$ \\

ChadaViT~\cite{bourriez2024chadavitchanneladaptive} &
$35.7{\tiny\,(0.2)}$ &
$34.3{\tiny\,(0.2)}$ &
$35.6{\tiny\,(0.1)}$ \\

OpenPhenom~\cite{rxrx_openphenom} &
$33.4{\tiny\,(0.3)}$ &
$31.5{\tiny\,(0.4)}$ &
$32.9{\tiny\,(0.4)}$ \\

\bottomrule
\end{tabular}%

\caption{Downstream multi-class classification on HPA~\cite{hpa_single_cell_classification} with frozen embedding models. Values represent the mean (standard deviation) across 5 folds, reported as percentages. The only model outperforming HASSL, Cellpaint-DINO~\cite{kim2025self}, likely benefits from exposure to organelle-targeted genetic perturbations and a similar immunofluorescent dye in JUMP-CP~\cite{jump_cp}. Nevertheless, HASSL remains close to the top-performing model and substantially improves over the other baselines, without having seen the modality or the data type.}
\label{tab:downstream_cls_hpa}
\end{table}

%On the HPA~\cite{hpa_single_cell_classification} single cell classification dataset, a model specifically tuned to HPA-like data, Cellpaint DINO~\cite{kim2025self} performs best (Tab.~\ref{tab:downstream_cls_hpa}). It was trained on JUMP-CP \cite{jump_cp}, which contains immunofluorescent dyes and genetic perturbations to organelles, resulting in significant carryover to the HPA task of organelle classification.

However, HASSL yields a performance very close to Cellpaint-DINO while outperforming every other model, including the baseline DINOv3~\cite{dinov3}, despite never seeing HPA during training. This indicates that encouraging morphology-driven structure improves transferability across modalities, even when the external label differs. \\
\noindent \textbf{Compute Overhead.}
Over 100 epochs, overhead is small: +43.6 min (+2.3\%) wall-clock and +4.44 GB (+10.2\%) VRAM vs. DINOv3 (32h01m/43.44 GB → 32h44m/47.88 GB) at identical GPU power and clocks, enabled by frozen teacher passes, $seg \rightarrow img$ reusing DINO activations, and checkpoint-bounded $seg\rightarrow seg$.

\section{Conclusion and Future Work}
We propose a hierarchy-aware self-supervised framework for learning representations from single-cell microscopy images. Our method adds a drop-in objective for self-distillation embedding models by constructing hierarchical prototypes via in-batch HDBSCAN and weighting their reliability using clustering confidence. Based on these prototypes, we perform hierarchy-consistent pair mining: positives favor ancestors and confident siblings, while negatives are scaled by hierarchical distance to avoid over-repelling closely related subtypes. A contrastive objective aligns samples with prototypes while enforcing cross-parent separation, and a double-teacher design yields more structured embedding spaces that better capture subtle morphological differences. Extensive experiments on clustering quality and downstream tasks demonstrate the effectiveness of our approach and motivate hierarchy-aware supervision for large-scale cellular representation learning and microscopy foundation models. The scope of our work is not limited to DINO-based architectures only. Segmentation embeddings can define pseudo-positive and pseudo-negative pairs for any SSL or class-guided objective (e.g., BYOL~\cite{byol}, Triplet Loss~\cite{FaceNet}, SimCLR~\cite{simclr}), while the HDBSCAN term is an additive batch-level loss, which can be used with any SSL setup. These variants, including replacing the hinge loss with InfoNCE~\cite{infonce}, can be explored in future work.

\subsubsection{\ackname} C.M. acknowledges support from the European Research Council (ERC; Grant Nos. 866411, 101113551, and 101213822), the High-tech Agenda Bayern, and the Deutsche Forschungsgemeinschaft (DFG, TRR359, Project No. 491676693). This work was supported by the de.NBI Cloud within the German Network for Bioinformatics Infrastructure (de.NBI) and ELIXIR-DE.

\bibliographystyle{splncs04}
\bibliography{main}

@String(CVPR= {IEEE Conf. Comput. Vis. Pattern Recog.})

@String(ICCV= {Int. Conf. Comput. Vis.})

@String(NIPS= {Adv. Neural Inform. Process. Syst.})

@String(ICLR = {Int. Conf. Learn. Represent.})

@String(AAAI = {AAAI})

@String(CVPR  = {CVPR})

@String(ICCV  = {ICCV})

@String(NIPS  = {NeurIPS})

@String(ICLR  = {ICLR})

@inproceedings{simclr,
  title={A simple framework for contrastive learning of visual representations},
  author={Chen, Ting and Kornblith, Simon and Norouzi, Mohammad and Hinton, Geoffrey},
  booktitle={ICML},
  year={2020}
}

@INPROCEEDINGS{FaceNet,
  author={Schroff, Florian and Kalenichenko, Dmitry and Philbin, James},
  booktitle={CVPR}, 
  title={FaceNet: A unified embedding for face recognition and clustering}, 
  year={2015},
  doi={10.1109/CVPR.2015.7298682}
}

@misc{kingma2013auto,
  title={Auto-encoding variational bayes},
  author={Kingma, Diederik P and Welling, Max and others},
  year={2013},
  publisher={Banff, Canada}
}

@article{van2017neural,
  title={Neural discrete representation learning},
  author={Van Den Oord, Aaron and Vinyals, Oriol and others},
  journal={Advances in neural information processing systems},
  year={2017}
}

@article{dosovitskiy2020image,
  title={An image is worth 16x16 words: Transformers for image recognition at scale},
  author={Dosovitskiy, Alexey and Beyer, Lucas and Kolesnikov, Alexander and Weissenborn, Dirk and Zhai, Xiaohua and Unterthiner, Thomas and Dehghani, Mostafa and Minderer, Matthias and Heigold, Georg and Gelly, Sylvain and others},
  journal={arXiv preprint arXiv:2010.11929},
  year={2020}
}

@inproceedings{khosla2020supervised,
  title     = {Supervised Contrastive Learning},
  author    = {Khosla, Prannay and Teterwak, Piotr and Wang, Chen and Sarna, Aaron and Tian, Yonglong and Isola, Phillip and Maschinot, Aaron and Liu, Ce and Krishnan, Dilip},
  booktitle = {NeurIPS},
  year      = {2020}
}

@inproceedings{haslum2024metadata,
  title     = {Metadata-guided Consistency Learning for High Content Images},
  author    = {Haslum, Johan Fredin and Matsoukas, Christos and Leuchowius, Karl-Johan and M{\"u}llers, Erik and Smith, Kevin},
  booktitle = {Medical Imaging with Deep Learning (MIDL), Proceedings of Machine Learning Research},
  volume    = {227},
  pages     = {918--936},
  year      = {2024},
  url       = {https://proceedings.mlr.press/v227/haslum24a.html}
}

@inproceedings{yao2024weakly,
  title={Weakly supervised set-consistency learning improves morphological profiling of single-cell images},
  author={Yao, Heming and Hanslovsky, Phil and Huetter, Jan-Christian and Hoeckendorf, Burkhard and Richmond, David},
  booktitle={Proceedings of the IEEE/CVF Conference on Computer Vision and Pattern Recognition},
  pages={6978--6987},
  year={2024}
}

@article{arevalo2024batch,
  title   = {Evaluating batch correction methods for image-based cell profiling},
  author  = {Arevalo, John and Su, Ellen and Ewald, Jessica D. and van Dijk, Robert and Carpenter, Anne E. and Singh, Shantanu},
  journal = {Nature Communications},
  volume  = {15},
  pages   = {6516},
  year    = {2024},
  doi     = {10.1038/s41467-024-50613-5}
}

@article{dinov2,
  title={Dinov2: Learning robust visual features without supervision},
  author={Oquab, Maxime and Darcet, Timoth{\'e}e and Moutakanni, Th{\'e}o and Vo, Huy and Szafraniec, Marc and Khalidov, Vasil and Fernandez, Pierre and Haziza, Daniel and Massa, Francisco and El-Nouby, Alaaeldin and others},
  journal={arXiv preprint arXiv:2304.07193},
  year={2023}
}

@article{dinov3,
  title = {{DINOv3}},
  author = {Sim{\'e}oni, Oriane and Vo, Huy V. and Seitzer, Maximilian and Baldassarre, Federico and Oquab, Maxime and Jose, Cijo and Khalidov, Vasil and Szafraniec, Marc and Yi, Seungeun and Ramamonjisoa, Micha{\"e}l and Massa, Francisco and Haziza, Daniel and Wehrstedt, Luca and Wang, Jianyuan and Darcet, Timoth{\'e}e and Moutakanni, Th{\'e}o and Sentana, Leonel and Roberts, Claire and Vedaldi, Andrea and Tolan, Jamie and Brandt, John and Couprie, Camille and Mairal, Julien and J{\'e}gou, Herv{\'e} and Labatut, Patrick and Bojanowski, Piotr},
  journal = {arXiv preprint arXiv:2508.10104},
  year = {2025},
  doi = {10.48550/arXiv.2508.10104}
}

@inproceedings{Guo2022HCSC,
  title={HCSC: Hierarchical Contrastive Selective Coding},
  author={Guo, Yuanfan and Xu, Minghao and Li, Jiawen and Ni, Bingbing and Zhu, Xuanyu and Sun, Zhenbang and Xu, Yi},
  booktitle={IEEE/CVF Conference on Computer Vision and Pattern Recognition (CVPR)},
  year={2022},
  url={https://openaccess.thecvf.com/content/CVPR2022/papers/Guo_HCSC_Hierarchical_Contrastive_Selective_Coding_CVPR_2022_paper.pdf}
}

@article{Xu2023SeedViews,
  title={Seed the Views: Hierarchical Semantic Alignment for Contrastive Representation Learning},
  author={Xu, Haohang and Zhang, Xiaopeng and Li, Hao and Xie, Lingxi and Dai, Wenrui and Xiong, Hongkai and Tian, Qi},
  journal={IEEE Transactions on Pattern Analysis and Machine Intelligence},
  volume={45},
  number={3},
  pages={3753--3767},
  year={2023},
  doi={10.1109/TPAMI.2022.3176690},
  url={https://arxiv.org/abs/2012.02733}
}

@article{Znalezniak2023CoHiClust,
  title={Contrastive Hierarchical Clustering},
  author={Znale{\'z}niak, Micha{\l} and Rola, Przemys{\l}aw and Kaszuba, Patryk and Tabor, Jacek and {\'S}mieja, Marek},
  journal={arXiv preprint arXiv:2303.03389},
  year={2023},
  url={https://arxiv.org/abs/2303.03389}
}

@inproceedings{He2020MoCo,
  title     = {Momentum Contrast for Unsupervised Visual Representation Learning},
  author    = {He, Kaiming and Fan, Haoqi and Wu, Yuxin and Xie, Saining and Girshick, Ross B.},
  booktitle = {Proceedings of the IEEE/CVF Conference on Computer Vision and Pattern Recognition (CVPR)},
  year      = {2020},
  url       = {https://arxiv.org/abs/1911.05722}
}

@inproceedings{Wu2018InstanceDiscrimination,
  title     = {Unsupervised Feature Learning via Non-Parametric Instance Discrimination},
  author    = {Wu, Zhirong and Xiong, Yuanjun and Yu, Stella X. and Lin, Dahua},
  booktitle = {Proceedings of the IEEE/CVF Conference on Computer Vision and Pattern Recognition (CVPR)},
  year      = {2018},
  url       = {https://arxiv.org/abs/1805.01978}
}

@InProceedings{HDBSCAN,
author="Campello, Ricardo J. G. B.
and Moulavi, Davoud
and Sander, Joerg",
editor="Pei, Jian
and Tseng, Vincent S.
and Cao, Longbing
and Motoda, Hiroshi
and Xu, Guandong",
title="Density-Based Clustering Based on Hierarchical Density Estimates",
booktitle="Advances in Knowledge Discovery and Data Mining",
year="2013",
publisher="Springer Berlin Heidelberg",
address="Berlin, Heidelberg",
pages="160--172",
isbn="978-3-642-37456-2"
}

@article{DEPTO2021101653,
title = {Automatic segmentation of blood cells from microscopic slides: A comparative analysis},
journal = {Tissue and Cell},
volume = {73},
pages = {101653},
year = {2021},
issn = {0040-8166},
doi = {https://doi.org/10.1016/j.tice.2021.101653},
url = {https://www.sciencedirect.com/science/article/pii/S0040816621001695},
author = {Deponker Sarker Depto and Shazidur Rahman and Md. Mekayel Hosen and Mst Shapna Akter and Tamanna Rahman Reme and Aimon Rahman and Hasib Zunair and M. Sohel Rahman and M.R.C. Mahdy}
}

@misc{graham2021coniccolonnucleiidentification,
      title={CoNIC: Colon Nuclei Identification and Counting Challenge 2022}, 
      author={Simon Graham and Mostafa Jahanifar and Quoc Dang Vu and Giorgos Hadjigeorghiou and Thomas Leech and David Snead and Shan E Ahmed Raza and Fayyaz Minhas and Nasir Rajpoot},
      year={2021},
      eprint={2111.14485},
      archivePrefix={arXiv},
      primaryClass={cs.CV},
      url={https://arxiv.org/abs/2111.14485}, 
}

@article{Vu2019Methods,
  title        = {Methods for Segmentation and Classification of Digital Microscopy Tissue Images},
  author       = {Vu, Quoc Dang and Graham, Simon and Kurc, Tahsin and To, Minh Nguyen Nhat and Shaban, Muhammad and Qaiser, Talha and Koohbanani, Navid Alemi and Khurram, Syed Ali and Kalpathy-Cramer, Jayashree and Zhao, Tianhao and Gupta, Rajarsi and Kwak, Jin Tae and Rajpoot, Nasir and Saltz, Joel and Farahani, Keyvan},
  journal      = {Frontiers in Bioengineering and Biotechnology},
  volume       = {7},
  pages        = {53},
  year         = {2019},
  doi          = {10.3389/fbioe.2019.00053},
  url          = {https://www.frontiersin.org/articles/10.3389/fbioe.2019.00053/full},
  publisher    = {Frontiers Media SA},
  month        = apr
}

@article{Naylor2019DistanceMap,
  title   = {Segmentation of Nuclei in Histopathology Images by Deep Regression of the Distance Map},
  author  = {Naylor, Peter and La{\'e}, Marick and Reyal, Fabien and Walter, Thomas},
  journal = {IEEE Transactions on Medical Imaging},
  year    = {2019},
  month   = feb,
  volume  = {38},
  number  = {2},
  pages   = {448--459},
  doi     = {10.1109/TMI.2018.2865709},
  url     = {https://doi.org/10.1109/TMI.2018.2865709}
}

@article{MAHBOD2021104349,
title = {CryoNuSeg: A dataset for nuclei instance segmentation of cryosectioned H\&E-stained histological images},
journal = {Computers in Biology and Medicine},
volume = {132},
pages = {104349},
year = {2021},
issn = {0010-4825},
doi = {https://doi.org/10.1016/j.compbiomed.2021.104349},
url = {https://www.sciencedirect.com/science/article/pii/S0010482521001438},
author = {Amirreza Mahbod and Gerald Schaefer and Benjamin Bancher and Christine Löw and Georg Dorffner and Rupert Ecker and Isabella Ellinger}
}

@article {Cellpose1,
	author = {Stringer, Carsen and Michaelos, Michalis and Pachitariu, Marius},
	title = {Cellpose: a generalist algorithm for cellular segmentation},
	elocation-id = {2020.02.02.931238},
	year = {2020},
	doi = {10.1101/2020.02.02.931238},
	publisher = {Cold Spring Harbor Laboratory},
	URL = {https://www.biorxiv.org/content/early/2020/02/03/2020.02.02.931238},
	eprint = {https://www.biorxiv.org/content/early/2020/02/03/2020.02.02.931238.full.pdf},
	journal = {bioRxiv}
}

@article {CellposeSAM,
	author = {Pachitariu, Marius and Rariden, Michael and Stringer, Carsen},
	title = {Cellpose-SAM: superhuman generalization for cellular segmentation},
	elocation-id = {2025.04.28.651001},
	year = {2025},
	doi = {10.1101/2025.04.28.651001},
	publisher = {Cold Spring Harbor Laboratory},
	URL = {https://www.biorxiv.org/content/early/2025/05/01/2025.04.28.651001},
	eprint = {https://www.biorxiv.org/content/early/2025/05/01/2025.04.28.651001.full.pdf},
	journal = {bioRxiv}
}

@misc{data-science-bowl-2018,
    author = {Allen Goodman and Anne Carpenter and Elizabeth Park and jlefman-nvidia and Josette BoozAllen and Kyle and Maggie and Nilofer and Peter Sedivec and Will Cukierski},
    title = {2018 Data Science Bowl },
    year = {2018},
    note = {Kaggle}
}

@misc{DynamicNuclearNetSegmentation_v1_0,
  title         = {DynamicNuclearNet Segmentation (v1.0)},
  author        = {{Van Valen Lab (Caltech)}},
  year          = {2023},
  note          = {DeepCell Datasets; non-commercial academic use (modified Apache license); accessed 2025-10-27},
  url           = {https://deepcell.readthedocs.io/en/master/data-gallery/dynamicnuclearnet.html},
  organization  = {DeepCell / Van Valen Lab}
}

@article{wang2024simultaneously,
  title={Simultaneously segmenting and classifying cell nuclei by using multi-task learning in multiplex immunohistochemical tissue microarray sections},
  author={Wang, Ranran and Qiu, Yusong and Hao, Xinyu and Jin, Shan and Gao, Junxiu and Qi, Heng and Xu, Qi and Zhang, Yong and Xu, Hongming},
  journal={Biomedical Signal Processing and Control},
  volume={93},
  pages={106143},
  year={2024},
  publisher={Elsevier}
}

@misc{Pfaendler2022_iPSC_Morpho,
  title     = {Morphologically annotated single-cell images of human induced pluripotent stem cells for deep learning},
  author    = {Pfaendler, Ramon},
  year      = {2022},
  month     = nov,
  publisher = {ETH Zurich},
  address   = {Zurich},
  doi       = {10.3929/ethz-b-000581447},
  url       = {https://www.research-collection.ethz.ch/entities/researchdata/b48aa89f-0c90-44e9-ad67-325ae37e3e89},
  urldate   = {2025-10-27},
  note      = {Data Collection; Creative Commons Attribution-ShareAlike 4.0 International}
}

@misc{naji_hussein_2023_8065174,
  author       = {Naji Hussein and
                  Büttner Reinhard and
                  Simon Adrian and
                  Eich Marie-Lisa and
                  Lohneis Philipp and
                  Bozek Katarzyna},
  title        = {LyNSeC: Lymphoma Nuclear Segmentation and
                   Classification
                  },
  month        = jun,
  year         = 2023,
  publisher    = {Zenodo},
  doi          = {10.5281/zenodo.8065174},
  url          = {https://doi.org/10.5281/zenodo.8065174},
}

@ARTICLE{9446924,
  author={Verma, Ruchika and Kumar, Neeraj and others},
  journal={IEEE Transactions on Medical Imaging}, 
  title={MoNuSAC2020: A Multi-Organ Nuclei Segmentation and Classification Challenge}, 
  year={2021},
  volume={40},
  number={12},
  pages={3413-3423},
  doi={10.1109/TMI.2021.3085712}}

@inproceedings{dino,
  title     = {Emerging Properties in Self-Supervised Vision Transformers},
  author    = {Caron, Mathilde and Misra, Ishan and Mairal, Julien and Goyal, Priya and Bojanowski, Piotr and Joulin, Armand},
  booktitle = ICCV,
  year      = {2021}
}

@ARTICLE{8880654,
  author={Kumar, Neeraj and Verma, Ruchika and Anand, Deepak and others},
  journal={IEEE Transactions on Medical Imaging}, 
  title={A Multi-Organ Nucleus Segmentation Challenge}, 
  year={2020},
  volume={39},
  number={5},
  pages={1380-1391},
  doi={10.1109/TMI.2019.2947628}}

@article{NeurIPS-CellSeg,
      title = {The Multi-modality Cell Segmentation Challenge: Towards Universal Solutions},
      author = {Jun Ma and Ronald Xie and Shamini Ayyadhury and Cheng Ge and Anubha Gupta and Ritu Gupta and Song Gu and Yao Zhang and Gihun Lee and Joonkee Kim and Wei Lou and Haofeng Li and Eric Upschulte and Timo Dickscheid and José Guilherme de Almeida and Yixin Wang and Lin Han and Xin Yang and Marco Labagnara and Vojislav Gligorovski and Maxime Scheder and Sahand Jamal Rahi and Carly Kempster and Alice Pollitt and Leon Espinosa and Tâm Mignot and Jan Moritz Middeke and Jan-Niklas Eckardt and Wangkai Li and Zhaoyang Li and Xiaochen Cai and Bizhe Bai and Noah F. Greenwald and David Van Valen and Erin Weisbart and Beth A. Cimini and Trevor Cheung and Oscar Brück and Gary D. Bader and Bo Wang},
      journal = {Nature Methods},
      volume={21},
      pages={1103–1113},
      year = {2024},
      doi = {https://doi.org/10.1038/s41592-024-02233-6}
  }

@article{mahbod2023nuinsseg,
  title={NuInsSeg: A Fully Annotated Dataset for Nuclei Instance Segmentation in H\&E-Stained Histological Images},
  author={Mahbod, Amirreza and Polak, Christine and Feldmann, Katharina and Khan, Rumsha and Gelles, Katharina and Dorffner, Georg and Woitek, Ramona and 
          Hatamikia, Sepideh and Ellinger, Isabella},
  journal={arXiv preprint arXiv:2308.01760},
  year={2023}
}

@article{Cutler2022,
  title   = {{Omnipose}: a high-precision morphology-independent solution for bacterial cell segmentation},
  author  = {Cutler, Kevin J. and Stringer, Carsen and Lo, Teresa W. and Rappez, Luca and Stroustrup, Nicholas and Brook Peterson, S. and Wiggins, Paul A. and Mougous, Joseph D.},
  journal = {Nature Methods},
  year    = {2022},
  month   = nov,
  volume  = {19},
  number  = {11},
  pages   = {1438--1448},
  issn    = {1548-7105},
  doi     = {10.1038/s41592-022-01639-4},
  url     = {https://doi.org/10.1038/s41592-022-01639-4}
}

@inproceedings{gamper2019pannuke,
  title={PanNuke: an open pan-cancer histology dataset for nuclei instance segmentation and classification},
  author={Gamper, Jevgenij and Koohbanani, Navid Alemi and Benes, Ksenija and Khuram, Ali and Rajpoot, Nasir},
  booktitle={European Congress on Digital Pathology},
  pages={11--19},
  year={2019},
  organization={Springer}
}

@article{gamper2020pannuke,
  title={PanNuke Dataset Extension, Insights and Baselines},
  author={Gamper, Jevgenij and Koohbanani, Navid Alemi and Graham, Simon and Jahanifar, Mostafa and Khurram, Syed Ali and Azam, Ayesha and Hewitt, Katherine and Rajpoot, Nasir},
  journal={arXiv preprint arXiv:2003.10778},
  year={2020}
}

@misc{Severin2021_PBMC_Morphology,
  title     = {Deep phenotyping reveals the molecular and health determinants of human immune cell morphology},
  author    = {Severin, Yannik},
  year      = {2021},
  month     = jun,
  publisher = {ETH Zurich},
  address   = {Zurich},
  doi       = {10.3929/ethz-b-000343106},
  url       = {https://www.research-collection.ethz.ch/entities/researchdata/d0feef10-453f-4b8d-b8e1-582ea2976f91},
  urldate   = {2025-10-27},
  note      = {Dataset; Creative Commons Attribution-ShareAlike 4.0 International (CC BY-SA 4.0)}
}

@article{Dietler2020,
  title = {A convolutional neural network segments yeast microscopy images with high accuracy.},
  author  = {Dietler, Nicola and Minder, Matthias and Gligorovski, Vojislav and Economou, Augoustina Maria and Joly, Denis Alain Henri Lucien and Sadeghi, Ahmad and Chan, Chun Hei Michael and Kozi{\'n}ski, Mateusz and Weigert, Martin and Bitbol, Anne-Florence and Rahi, Sahand Jamal},
  journal = {Nature Communications},
  year    = {2020},
  month   = nov,
  volume  = {11},
  number  = {1},
  pages   = {5723},
  issn    = {2041-1723},
  doi     = {10.1038/s41467-020-19557-4},
  url     = {https://doi.org/10.1038/s41467-020-19557-4}
}

@misc{DeepCell_TissueNet_1_1,
  title        = {TissueNet: training dataset for nuclear and whole-cell segmentation},
  author       = {{Van Valen Lab (Caltech)}},
  year         = {2022},
  month        = apr,
  version      = {1.1},
  note         = {DeepCell datasets documentation; modified Apache license (non-commercial academic use). Accessed 2025-10-27.}
}

@misc{sartorius-cell-instance-segmentation,
    author = {Addison Howard and Ashley Chow and CorporateResearchSartorius and Maria Ca and Phil Culliton and Tim Jackson},
    title = {Sartorius - Cell Instance Segmentation},
    year = {2021},
    note = {Kaggle}
}

@article{infonce,
  title   = {Representation Learning with Contrastive Predictive Coding},
  author  = {van den Oord, Aaron and Li, Yazhe and Vinyals, Oriol},
  journal = {arXiv preprint arXiv:1807.03748},
  year    = {2018}
}

@article{Huang2023,
  author    = {Huang, Shih-Cheng and Pareek, Anuj and Jensen, Malte and Lungren, Matthew P. and Yeung, Serena and Chaudhari, Akshay S.},
  title     = {Self-supervised learning for medical image classification: a systematic review and implementation guidelines},
  journal   = {npj Digital Medicine},
  volume    = {6},
  number    = {1},
  pages     = {74},
  year      = {2023},
  month     = {Apr},
  doi       = {10.1038/s41746-023-00811-0},
}

@article{Bendidi2024,
  author    = {Bendidi, Ihab and Bardes, Adrien and Cohen, Ethan and Lamiable, Alexis and Bollot, Guillaume and Genovesio, Auguste},
  title     = {Exploring self-supervised learning biases for microscopy image representation},
  journal   = {Biological Imaging},
  volume    = {4},
  pages     = {e12},
  year      = {2024},
  month     = {Nov},
  doi       = {10.1017/S2633903X2400014X},
}

@article{Kobayashi2022,
  author    = {Kobayashi, Hirofumi and Cheveralls, Keith C. and Leonetti, Manuel D. and Royer, Loic A.},
  title     = {Self-supervised deep learning encodes high-resolution features of protein subcellular localization},
  journal   = {Nature Methods},
  volume    = {19},
  number    = {8},
  pages     = {995--1003},
  year      = {2022},
  month     = {Aug},
  doi       = {10.1038/s41592-022-01541-z},
}

@INPROCEEDINGS{imagenet,
  author={Deng, Jia and Dong, Wei and Socher, Richard and Li, Li-Jia and Kai Li and Li Fei-Fei},
  booktitle={2009 IEEE Conference on Computer Vision and Pattern Recognition}, 
  title={ImageNet: A large-scale hierarchical image database}, 
  year={2009},
  volume={},
  number={},
  pages={248-255},
  doi={10.1109/CVPR.2009.5206848}}

@inproceedings{SWAV,
 author = {Caron, Mathilde and Misra, Ishan and Mairal, Julien and Goyal, Priya and Bojanowski, Piotr and Joulin, Armand},
 booktitle = {Advances in Neural Information Processing Systems},
 editor = {H. Larochelle and M. Ranzato and R. Hadsell and M.F. Balcan and H. Lin},
 pages = {9912--9924},
 publisher = {Curran Associates, Inc.},
 title = {Unsupervised Learning of Visual Features by Contrasting Cluster Assignments},
 url = {https://proceedings.neurips.cc/paper_files/paper/2020/file/70feb62b69f16e0238f741fab228fec2-Paper.pdf},
 volume = {33},
 year = {2020}
}

@article{Yang2023MedMNISTv2,
  title   = {MedMNIST v2 -- A large-scale lightweight benchmark for 2D and 3D biomedical image classification},
  author  = {Yang, Jiancheng and Shi, Rui and Wei, Donglai and Liu, Zequan and Zhao, Lin and Ke, Bilian and Pfister, Hanspeter and Ni, Bingbing},
  journal = {Scientific Data},
  year    = {2023},
  month   = jan,
  volume  = {10},
  number  = {1},
  pages   = {41},
  doi     = {10.1038/s41597-022-01721-8},
  url     = {https://doi.org/10.1038/s41597-022-01721-8},
  issn    = {2052-4463}
}

@misc{bourriez2024chadavitchanneladaptive,
      title={ChAda-ViT : Channel Adaptive Attention for Joint Representation Learning of Heterogeneous Microscopy Images}, 
      author={Nicolas Bourriez and Ihab Bendidi and Ethan Cohen and Gabriel Watkinson and Maxime Sanchez and Guillaume Bollot and Auguste Genovesio},
      year={2024},
      eprint={2311.15264},
      archivePrefix={arXiv},
      primaryClass={cs.CV},
      url={https://arxiv.org/abs/2311.15264}, 
}

@inproceedings{kraus2024masked,
  title={Masked Autoencoders for Microscopy are Scalable Learners of Cellular Biology},
  author={Kraus, Oren and Kenyon-Dean, Kian and Saberian, Saber and Fallah, Maryam and McLean, Peter and Leung, Jess and Sharma, Vasudev and Khan, Ayla and Balakrishnan, Jia and Celik, Safiye and others},
  booktitle={Proceedings of the IEEE/CVF Conference on Computer Vision and Pattern Recognition},
  pages={11757--11768},
  year={2024}
}

@misc{rxrx_openphenom,
  author       = {{Recursion Pharmaceuticals}},
  title        = {{OpenPhenom}: Groundbreaking publicly accessible foundation models for microscopy data},
  year         = {2025},
  note         = {Accessed October 31, 2025},
  urldate      = {2025-10-31}
}

@article{pfaendler2023self,
  title={Self-supervised vision transformers accurately decode cellular state heterogeneity},
  author={Pfaendler, Ramon and Hanimann, Jacob and Lee, Sohyon and Snijder, Berend},
  journal={Biorxiv},
  pages={2023--01},
  year={2023},
  publisher={Cold Spring Harbor Laboratory}
}

@article{gupta2024subcell,
  title={SubCell: Vision foundation models for microscopy capture single-cell biology},
  author={Gupta, Ankit and Wefers, Zoe and Kahnert, Konstantin and Hansen, Jan N and Leineweber, Will and Cesnik, Anthony and Lu, Dan and Axelsson, Ulrika and Ballllosera Navarro, Frederic and Karaletsos, Theofanis and others},
  journal={bioRxiv},
  pages={2024--12},
  year={2024},
  publisher={Cold Spring Harbor Laboratory}
}

@article{kim2025self,
  title={Self-supervision advances morphological profiling by unlocking powerful image representations},
  author={Kim, Vladislav and Adaloglou, Nikolaos and Osterland, Marc and Morelli, Flavio M and Halawa, Marah and K{\"o}nig, Tim and Gnutt, David and Marin Zapata, Paula A},
  journal={Scientific Reports},
  volume={15},
  number={1},
  pages={4876},
  year={2025},
  publisher={Nature Publishing Group UK London}
}

@inproceedings{Li2021PCL,
  author    = {Junnan Li and Pan Zhou and Caiming Xiong and Steven C. H. Hoi},
  title     = {Prototypical Contrastive Learning of Unsupervised Representations},
  booktitle = {9th International Conference on Learning Representations (ICLR)},
  year      = {2021}
}

@inproceedings{byol,
author = {Grill, Jean-Bastien and Strub, Florian and Altch\'{e}, Florent and Tallec, Corentin and Richemond, Pierre H. and Buchatskaya, Elena and Doersch, Carl and Pires, Bernardo Avila and Guo, Zhaohan Daniel and Azar, Mohammad Gheshlaghi and Piot, Bilal and Kavukcuoglu, Koray and Munos, R\'{e}mi and Valko, Michal},
title = {Bootstrap your own latent a new approach to self-supervised learning},
year = {2020},
isbn = {9781713829546},
publisher = {Curran Associates Inc.},
address = {Red Hook, NY, USA},
booktitle = {Proceedings of the 34th International Conference on Neural Information Processing Systems},
articleno = {1786},
numpages = {14},
location = {Vancouver, BC, Canada},
series = {NIPS '20}
}

@article{maaten2008visualizing,
  title={Visualizing data using t-SNE},
  author={Maaten, Laurens van der and Hinton, Geoffrey},
  journal={Journal of machine learning research},
  volume={9},
  number={Nov},
  pages={2579--2605},
  year={2008}
}

@article {jump_cp,
	author = {Chandrasekaran, Srinivas Niranj and Ackerman, Jeanelle and Alix, Eric and Ando, D. Michael and Arevalo, John and Bennion, Melissa and Boisseau, Nicolas and Borowa, Adriana and Boyd, Justin D. and Brino, Laurent and others},
	title = {JUMP Cell Painting dataset: morphological impact of 136,000 chemical and genetic perturbations},
	elocation-id = {2023.03.23.534023},
	year = {2023},
	doi = {10.1101/2023.03.23.534023},
	publisher = {Cold Spring Harbor Laboratory},
	URL = {https://www.biorxiv.org/content/early/2023/03/24/2023.03.23.534023},
	eprint = {https://www.biorxiv.org/content/early/2023/03/24/2023.03.23.534023.full.pdf},
	journal = {bioRxiv}
}

@misc{hpa_single_cell_classification,
    author = {Casper Winsnes and Emma Lundberg and Maggie and Phil Culliton and Trang Le and UAxelsson and Wei Ouyang},
    title = {Human Protein Atlas - Single Cell Classification},
    year = {2021},
    note = {Kaggle}
}

@misc{allen_cell_drug_perturbation_2018,
  author       = {{Allen Institute for Cell Science}},
  title        = {Drug perturbation pilot study},
  year         = {2018},
  howpublished = {[dataset]},
  url          = {https://www.allencell.org/drug-perturbation-pilot.html},
  note         = {Accessed 2026-02-24}
}

@inproceedings{Actual-DBSCAN,
author = {Ester, Martin and Kriegel, Hans-Peter and Sander, J\"{o}rg and Xu, Xiaowei},
title = {A density-based algorithm for discovering clusters in large spatial databases with noise},
year = {1996},
publisher = {AAAI Press},
booktitle = {Proceedings of the Second International Conference on Knowledge Discovery and Data Mining},
pages = {226–231},
numpages = {6},
location = {Portland, Oregon},
series = {KDD'96}
}

@article{AMI,
author = {Vinh, Nguyen Xuan and Epps, Julien and Bailey, James},
title = {Information Theoretic Measures for Clusterings Comparison: Variants, Properties, Normalization and Correction for Chance},
year = {2010},
issue_date = {3/1/2010},
publisher = {JMLR.org},
volume = {11},
issn = {1532-4435},
journal = {J. Mach. Learn. Res.},
month = dec,
pages = {2837–2854},
numpages = {18}
}

\appendix
\clearpage
\setcounter{page}{1}
\maketitlesupplementary
\section{Loss Calculation Algorithms}
To aid in the understanding of the code, Algs.~\ref{algo:dt} and \ref{algo:hdb} provide the pseudocode for the Double-Teacher DINO distillation and the HDBSCAN hierarchy-aware contrastive loss, i.e.\ the two components of our objective. All variables follow the notation introduced in Sec. 3.1 and 3.2.

\begin{algorithm}[H]
\footnotesize
\caption{Double-Teacher DINO Distillation}
\begin{algorithmic}[1]
\Require Batch $\mathcal{B} = \{x_1, \ldots, x_B\}$ with segmentation masks $\{m_1, \ldots, m_B\}$
\Require Student $S$, Image teacher $T_{\text{img}}$, Segmentation teacher $T_{\text{seg}}$
\Require Temperature $\tau$, mixing weight $\gamma \in [0,1]$
\State \textbf{// Generate Multi-Crop Views}
\For{each image $x_i \in \mathcal{B}$}
    \State Generate global crops: $\{x_i^{g_1}, x_i^{g_2}\}$, local crops: $\{x_i^{\ell_1}, \ldots, x_i^{\ell_L}\}$
    \State Generate segmentation crops: $\{m_i^{g_1}, m_i^{g_2}\}$, $\{m_i^{\ell_1}, \ldots, m_i^{\ell_L}\}$
\EndFor
\State $V \gets \{$all student views$\}$, $V_t \gets \{$global crops only$\}$
\State \textbf{// Teacher Targets with Sinkhorn-Knopp Centering}
\For{each global view $v_t \in V_t$}
    \State $z_{T,\text{img}}^{(v_t)} \gets T_{\text{img}}(x^{(v_t)})$; \quad $q_{\text{img}}^{(v_t)} \gets \text{SK}(z_{T,\text{img}}^{(v_t)}; \tau)$
    \State $z_{T,\text{seg}}^{(v_t)} \gets T_{\text{seg}}(m^{(v_t)})$; \quad $q_{\text{seg}}^{(v_t)} \gets \text{SK}(z_{T,\text{seg}}^{(v_t)}; \tau)$
\EndFor
\State \textbf{// Student Predictions}
\For{each view $v \in V$}
    \State $p_{S,\text{img}}^{(v)} \gets S_{\text{img}}(v)$; \quad $p_{S,\text{seg}}^{(v)} \gets S_{\text{seg}}(v)$
\EndFor
\State \textbf{// Pooled Cross-Entropy Loss}
\State Define: $\Phi(q, \{p\}) = \frac{1}{|V_t|(|V|-1)} \sum_{v_t \in V_t} \sum_{\substack{v_s \in V \\ v_s \neq v_t}} \text{CE}(q^{(v_t)}, p^{(v_s)})$
\State $\mathcal{L}_{\text{DINO}} \gets \Phi(q_{\text{img}}, \{p_{S,\text{img}}^{(v)}\}_{v \in V})$
\State $\mathcal{L}_{\text{seg}\to\text{seg}} \gets \Phi(q_{\text{seg}}, \{p_{S,\text{seg}}^{(v)}\}_{v \in V})$
\State $\mathcal{L}_{\text{img}\to\text{seg}} \gets \Phi(q_{\text{seg}}, \{p_{S,\text{img}}^{(v)}\}_{v \in V})$
\State $\mathcal{L}_{\text{DoubleTeacher}} \gets (1-\gamma) \mathcal{L}_{\text{DINO}} + \gamma (\mathcal{L}_{\text{seg}\to\text{seg}} + \mathcal{L}_{\text{img}\to\text{seg}})$
\State \Return $\mathcal{L}_{\text{DoubleTeacher}}$
\end{algorithmic}
\label{algo:dt}
\end{algorithm}

\begin{algorithm}[H]
\footnotesize
\caption{HDBSCAN Hierarchical Contrastive Loss}
\begin{algorithmic}[1]
\Require L2-normalized embeddings $\{\hat{x}_1, \ldots, \hat{x}_B\}$ from student model
\Require Margin $m$, stability epsilon $\varepsilon$
\State \textbf{// Hierarchical Clustering}
\State $\mathcal{T} \gets \text{HDBSCAN}(\{\hat{x}_i\}_{i=1}^B, \text{min\_cluster\_size}=2)$ \Comment{Condensed tree}
\For{each cluster node $c \in \mathcal{T}$}
    \State $\mathcal{C}_c \gets \{$member indices$\}$; \quad $\lambda_c \gets$ stability of $c$
    \State $\mu_c \gets \text{norm}\!\left(\frac{1}{|\mathcal{C}_c|}\sum_{j \in \mathcal{C}_c} \hat{x}_j\right)$ \Comment{Cluster prototype}
\EndFor
\State \textbf{// Mine Positives and Negatives}
\State $\mathcal{I} \gets \{$non-noise points with $\geq 1$ positive and $\geq 1$ negative$\}$
\For{each anchor $i \in \mathcal{I}$}
    \State $\mathcal{P}(i) \gets \{c_{i0}, \ldots, c_{iK}\}$ \Comment{Ancestor path: leaf $\to$ root}
    \State $P_i \gets \{\mu_{c_{ik}}\}_{k=0}^K$ \Comment{Positive prototypes}
    \State $\mathcal{S}(i) \gets \bigcup_{k=0}^{K-1} \big(\text{Children}(c_{i,k+1}) \setminus \{c_{ik}\}\big)$ \Comment{Sibling clusters}
    \State $N_i \gets \{\mu_c\}_{c \in \mathcal{S}(i)}$ \Comment{Negative prototypes}
\EndFor
\State \textbf{// Compute Stability Weights as shown in Section 3.2.3}
\State $\lambda_{\min}, \lambda_{\max} \gets \min_c \lambda_c, \max_c \lambda_c$
\For{each anchor $i \in \mathcal{I}$}
    \For{$k = 0$ to $K$} \Comment{Positive weights}
        \State $\alpha_{ik} \gets \frac{\max\!\big(\varepsilon, \frac{\lambda_{c_{ik}} - \lambda_{\min}}{\lambda_{\max} - \lambda_{\min}}\big)}{\sum_{u=0}^{K} \max\!\big(\varepsilon, \frac{\lambda_{c_{iu}} - \lambda_{\min}}{\lambda_{\max} - \lambda_{\min}}\big)}$
    \EndFor
    \For{each $c \in \mathcal{S}(i)$} \Comment{Negative weights (inverted)}
        \State $\beta_{ic} \gets \frac{\max\!\big(\varepsilon, \frac{\lambda_{\max} - \lambda_c}{\lambda_{\max} - \lambda_{\min}}\big)}{\sum_{c' \in \mathcal{S}(i)} \max\!\big(\varepsilon, \frac{\lambda_{\max} - \lambda_{c'}}{\lambda_{\max} - \lambda_{\min}}\big)}$
    \EndFor
\EndFor
\State \textbf{// Weighted Hinge Contrastive Loss}
\For{each anchor $i \in \mathcal{I}$}
    \State $s_{\text{ap}}(i) \gets \sum_{k=0}^{K} \alpha_{ik} \cdot (\hat{x}_i^\top \mu_{c_{ik}})$ \Comment{Weighted positive similarity}
    \State $s_{\text{an}}(i) \gets \sum_{c \in \mathcal{S}(i)} \beta_{ic} \cdot (\hat{x}_i^\top \mu_c)$ \Comment{Weighted negative similarity}
    \State $L_i \gets \big[m + s_{\text{an}}(i) - s_{\text{ap}}(i)\big]_+$ \Comment{Hinge loss}
\EndFor
\State $\mathcal{L}_{\text{HDBSCAN}} \gets \frac{1}{|\mathcal{I}|} \sum_{i \in \mathcal{I}} L_i$
\State \Return $\mathcal{L}_{\text{HDBSCAN}}$
\end{algorithmic}
\label{algo:hdb}
\end{algorithm}

\section{Dataset Description}

Table \ref{Supp:datasets} summarizes the number of single-cells and the key properties of each dataset used in our work.

% in preamble (once):
% \usepackage{array}
% \newcolumntype{P}[1]{>{\centering\arraybackslash}p{#1}}
\begin{table}[H]
\centering
\scriptsize
\setlength{\tabcolsep}{3pt}
\renewcommand{\arraystretch}{0.95}
\caption{Overview of 20 nucleus/cell-segmentation datasets, including image modality, cell types, and imaging context. Datasets without annotated cell types are marked as `U'. For datasets marked as `U', we use the dataset name as the cell-type label.}
\label{Supp:datasets}
\begin{tabularx}{\textwidth}{>{\raggedright\arraybackslash}p{2.6cm} >{\raggedright\arraybackslash}p{1.5cm} >{\centering\arraybackslash}p{0.8cm} >{\raggedright\arraybackslash}p{2.2cm} Y}
\hline
\textbf{Name} & \textbf{Number of Cells} & \textbf{Labels} & \textbf{Modality} & \textbf{Cell Types} \\
\hline
BCCD \cite{DEPTO2021101653} & 90,813 & U & Brightfield (blood smear) & Blood cells \\
CoNIC \cite{graham2021coniccolonnucleiidentification} & 7,696 &  & H\&E histology & Colon epithelial, stromal, immune, neutrophils, eosinophils \\
CPM 15+17 and TNBC \cite{Vu2019Methods,Naylor2019DistanceMap} & 11,624 &  & H\&E histology & Tumor, stromal, immune cells \\
CryoNuSeg \cite{MAHBOD2021104349} & 2,273 &  & H\&E frozen sections & 10-organ nuclei \\
Cyto and Cyto2 \cite{Cellpose1,CellposeSAM} & 71,783 & U & Mixed microscopy & Mixed cultured cells \\
Data Science Bowl 2018 \cite{data-science-bowl-2018} & 14,902 & U & Mixed (IF, BF) & Mixed species nuclei \\
Dynamic Nuclear Net \cite{DynamicNuclearNetSegmentation_v1_0} & 347,572 & U & Fluorescence (live-cell) & Cultured human nuclei \\
TissueNet \cite{DeepCell_TissueNet_1_1} & 866,884 & U & Multiplex IF & Multiple tissue cell types \\
IHC TMA \cite{wang2024simultaneously} & 7,154 &  & Multiplex IHC (TMA) & Tumor, immune nuclei \\
iPSC \cite{Pfaendler2022_iPSC_Morpho} & 35,308 &  & Multichannel microscopy & iPSCs \\
LynSec \cite{naji_hussein_2023_8065174} & 70,676 &  & H\&E histology & Lymphoma cells (DLBCL) \\
MoNuSAC \cite{9446924} & 28,744 &  & H\&E histology & Neoplastic, lymphocyte, macrophage, neutrophil \\
MoNuSeg \cite{8880654} & 16,031 & U & H\&E histology & Mixed tumor and stromal nuclei \\
NeurIPS 2022 Cell-Seg \cite{NeurIPS-CellSeg} & 98,465 & U & Mixed microscopy & Cultured and tissue cells \\
NuInsSeg \cite{mahbod2023nuinsseg} & 25,293 &  & H\&E histology & 31-organ nuclei \\
Omnipose \cite{Cutler2022} & 37,038 &  & Phase contrast, fluorescence & Bacterial and other cells \\
PanNuke \cite{gamper2019pannuke,gamper2020pannuke} & 104,594 &  & H\&E histology & Tumor, immune, epithelial, stromal, dead \\
Phenoplex \cite{Severin2021_PBMC_Morphology} & 497,577 &  & Fluorescent confocal & PBMCs \\
Sartorius Challenge \cite{sartorius-cell-instance-segmentation} & 34,621 &  & Phase contrast & Cortical neurons, astrocytes, SH-SY5Y \\
YeaZ \cite{Dietler2020} & 21,784 & U & Phase contrast, brightfield & \textit{S. cerevisiae} \\
\hline
\end{tabularx}
\end{table}

\section{Hierarchy in the dataset}

The hierarchy in the dataset is shown in Fig.~\ref{fig:hierarchy}.

\begin{figure}[H]
    \centering
    \includegraphics[width=1.05\textwidth]{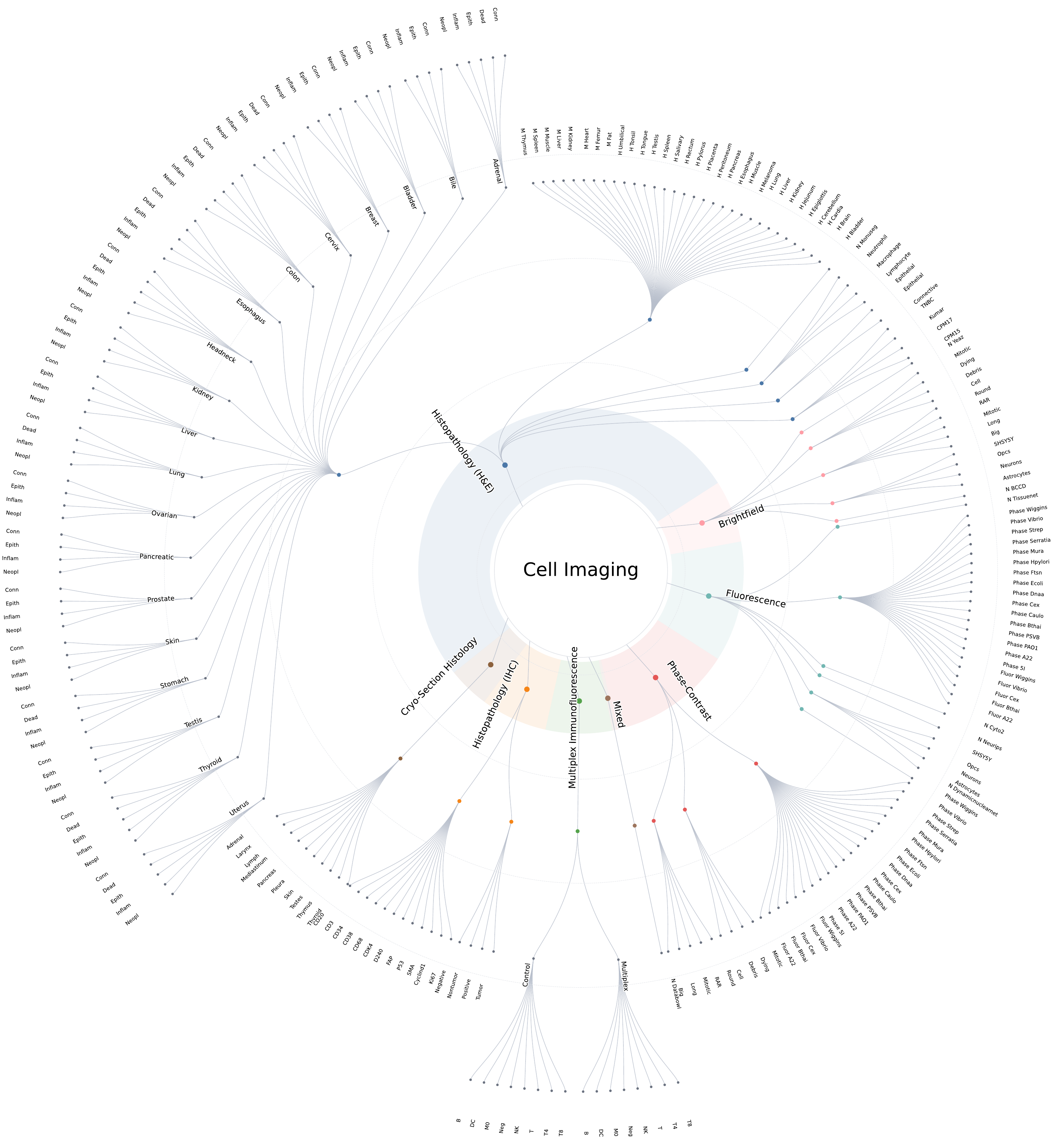}
    \caption{Radial Diagram showing the modalities and cell classes available in our dataset. The dataset exhibits multi-level hierarchies, including modalities, dataset/collection method, cell types, and subtypes.}
    \label{fig:hierarchy}
\end{figure}

\section{Result Comparison}

\begin{figure}[H]
    \centering
    \includegraphics[width=\linewidth]{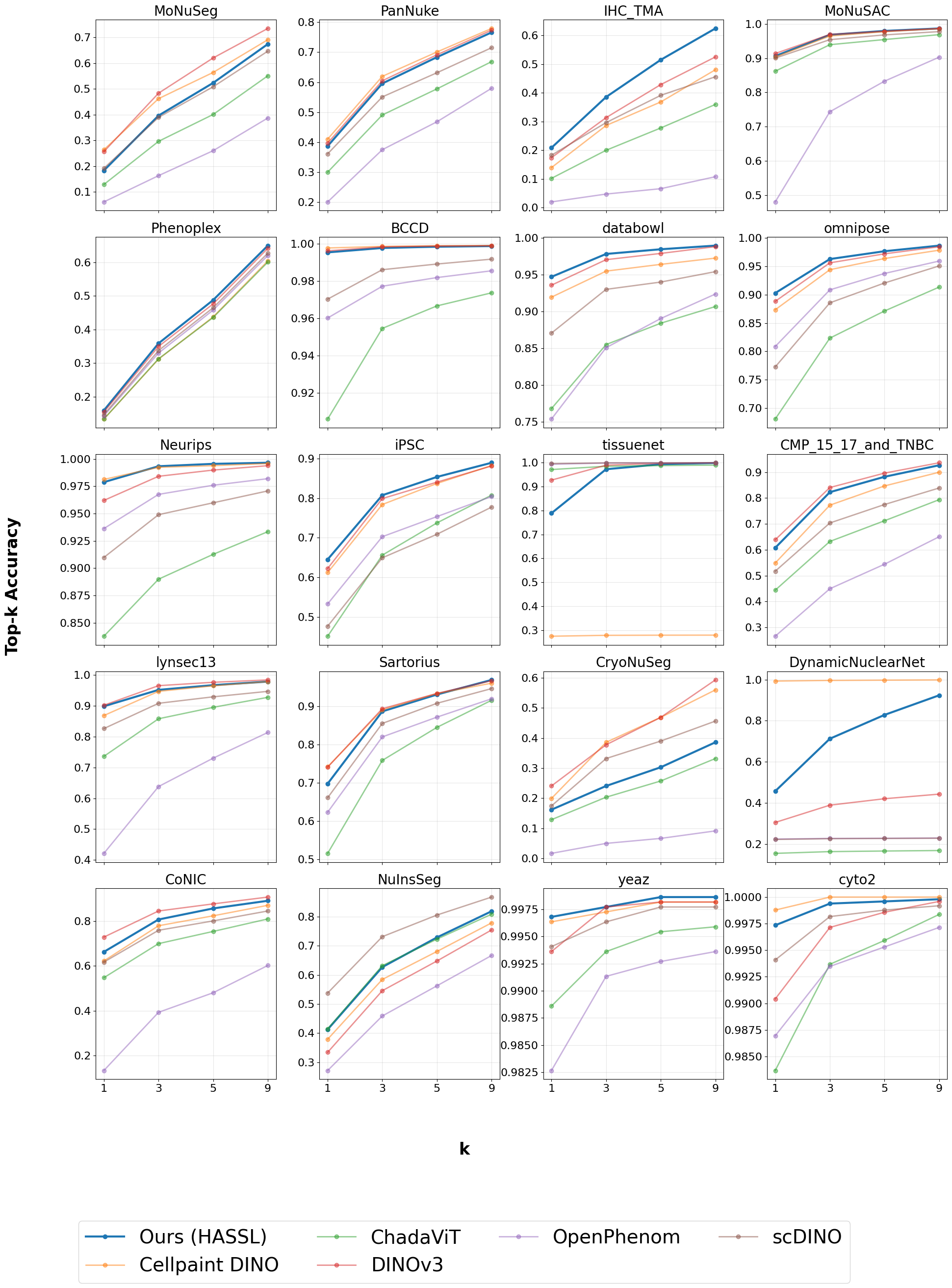}
    \caption{Comparison of the top-K retrieval results on all the individual datasets in our collection. Our model consistently returns high scores and beats the baselines in the majority of the cases, leading to a high average accuracy score across the test dataset.}
    \label{fig:perdataset}
\end{figure}

The top-K accuracy for each dataset is shown in Fig.~\ref{fig:perdataset}.

\section{Additional Results}
The following Table \ref{tab:modality_retrieval} shows the retrieval results aggregated over modality.
\begin{table}[H]
\centering

\resizebox{\linewidth}{!}{%
\begin{tabular}{lcccccccccccc}
\toprule
Model & \multicolumn{3}{c}{Fluor.} & \multicolumn{3}{c}{H\&E} & \multicolumn{3}{c}{Multiplex} & \multicolumn{3}{c}{Multichannel} \\
\cmidrule(lr){2-4}
\cmidrule(lr){5-7}
\cmidrule(lr){8-10}
\cmidrule(lr){11-13}
 & Top-1 & Top-9 & mAP & Top-1 & Top-9 & mAP & Top-1 & Top-9 & mAP & Top-1 & Top-9 & mAP \\
\midrule
CellPaint DINO& \textbf{50.3} & \textbf{77.5} & \textbf{56.3} & 58.3 & \textbf{85.5} & 63.2 & 14.0 & 48.1 & 21.6 & 61.2 & 88.2 & 66.0 \\
scDINO & 20.3 & 48.4 & 26.5 & 56.1 & 82.2 & 60.4 & 18.3 & 45.6 & 23.4 & 47.7 & 77.7 & 53.1 \\
OpenPhenom & 20.3 & 48.0 & 26.4 & 28.7 & 67.3 & 36.5 & 2.0 & 10.8 & 4.1 & 53.3 & 80.4 & 57.8 \\
\midrule
DINOv3 & 26.8 & 69.3 & 36.1 & 57.5 & 84.9 & 62.4 & 16.7 & 52.2 & 25.2 & 63.1 & 88.9 & 67.9 \\
\textbf{HASSL (Ours)} & 30.8 & 77.2 & 41.3 & \textbf{58.4} & 85.3 & \textbf{63.2} & \textbf{20.9} & \textbf{62.4} & \textbf{30.8} & \textbf{64.5} & \textbf{89.9} & \textbf{68.1} \\
\bottomrule
\end{tabular}}
\caption{KNN retrieval performance aggregated by imaging modality. Best per column in \textbf{bold}.}
\label{tab:modality_retrieval}
\end{table}
\noindent
The following Table \ref{tab:cross_modal} shows the results computed on AICS (unseen), measuring cross-modal retrieval. Here, we have used the embeddings from the Brightfield channel to retrieve the drug-aligned embedding from the Fluorescent channel. Here, HASSL outperforms all the baselines in this case.
\begin{table}
\centering
\label{tab:cross_modal_bf2fl}
\resizebox{\linewidth}{!}{%
\begin{tabular}{l cc cc cc cc c}
\toprule
 & \multicolumn{2}{c}{$K{=}1$} & \multicolumn{2}{c}{$K{=}3$} & \multicolumn{2}{c}{$K{=}5$} & \multicolumn{2}{c}{$K{=}9$} & \\
\cmidrule(lr){2-3}
\cmidrule(lr){4-5}
\cmidrule(lr){6-7}
\cmidrule(lr){8-9}
Method & Acc & Prec & Acc & Prec & Acc & Prec & Acc & Prec & mAP \\
\midrule
OpenPhenom & 35.5 & 35.5 & 45.4 & 34.0 & 46.9 & 33.9 & 49.6 & 34.0 & 40.1 \\
scDINO & 37.2 & 37.2 & 50.0 & 37.2 & 70.1 & 37.1 & 86.8 & 37.2 & 47.1 \\
\midrule
DINOv3 Baseline & 43.7 & 43.7 & 58.0 & 42.8 & 66.5 & 42.6 & 77.6 & 42.6 & 50.2 \\
HASSL (w/o DT) & 52.9 & 52.9 & 75.3 & 48.6 & 84.1 & 47.7 & 91.5 & 46.1 & 59.5 \\
HASSL (w/o HDBSCAN) & 53.8 & 53.8 & 75.1 & 48.7 & 83.7 & 47.2 & 91.5 & 45.9 & 59.4 \\
HASSL (ours)& \textbf{54.7} & \textbf{54.7} & \textbf{77.5} & \textbf{49.5} & \textbf{85.7} & \textbf{47.9} & \textbf{92.2} & \textbf{46.6} & \textbf{59.7} \\
\bottomrule
\end{tabular}}
\caption{Cross-modal retrieval (brightfield $\rightarrow$ fluorescence) on AICS. CellPaint-DINO's weights are no longer public (skipped).}
\label{tab:cross_modal}
\end{table}

%\begin{figure*}[t]
%    \centering
%    \includegraphics[width=0.85\textwidth]{Figures/comparison_visualization.png}
%    \caption{Example of Cell Class prediction task with HASSL (ours) and a DINOv3 (baseline). The images show that across modalities, HASSL is capable to capturing subtle morphological differences to differentiate cell classes better.}
    %\label{fig:enter-label}
%\end{figure*}

% \section{Rationale}
% \label{sec:rationale}
% % 
% Having the supplementary compiled together with the main paper means that:
% % 
% \begin{itemize}
% \item The supplementary can back-reference sections of the main paper, for example, we can refer to \cref{sec:intro};
% \item The main paper can forward reference sub-sections within the supplementary explicitly (e.g. referring to a particular experiment); 
% \item When submitted to arXiv, the supplementary will already included at the end of the paper.
% \end{itemize}
% % 
% To split the supplementary pages from the main paper, you can use \href{https://support.apple.com/en-ca/guide/preview/prvw11793/mac#:~:text=Delete%20a%20page%20from%20a,or%20choose%20Edit%20%3E%20Delete).}{Preview (on macOS)}, \href{https://www.adobe.com/acrobat/how-to/delete-pages-from-pdf.html#:~:text=Choose%20%E2%80%9CTools%E2%80%9D%20%3E%20%E2%80%9COrganize,or%20pages%20from%20the%20file.}{Adobe Acrobat} (on all OSs), as well as \href{https://superuser.com/questions/517986/is-it-possible-to-delete-some-pages-of-a-pdf-document}{command line tools}.

% \bibliographystyle{ieeenat_fullname}
% \bibliography{main}

\end{document}